\documentclass{ecai} 

\usepackage[utf8]{inputenc}

\usepackage{soul}
\usepackage{graphicx, array, blindtext}
\usepackage[linesnumbered]{algorithm2e}
\usepackage{url}
\usepackage{booktabs}
\usepackage{amsfonts}
\usepackage{subcaption}
\usepackage{mathbbol}
\usepackage{multirow}
\usepackage{amsmath}
\usepackage{pbox}
\usepackage{arydshln}
\usepackage[numbers]{natbib}
\usepackage[capitalise]{cleveref} 
\usepackage[fleqn]{nccmath}
\usepackage{graphicx}
\usepackage{amssymb}
\usepackage{multirow}

\def\defterm#1{\textbf{#1}}
\def\bs#1{\boldsymbol{#1}}
\def\set#1{\bs{#1}}

\newcommand{\tr}{\mathit{tr}}
\newcommand{\te}{\mathit{te}}

\newcommand{\test}[1]{\restrict{#1}{\te}}
\newcommand{\train}[1]{\restrict{#1}{\tr}}

\newcommand{\tar}{\Yvariable}

\newcommand{\target}[1]{#1^{\tar}}

\newcommand{\ev}{\set{E}}
\newcommand{\evinstance}{\set{e}}
\newcommand{\evidence}[1]{#1^{\ev}}

\newcommand{\loss}{\mathcal{L}}

\newcommand{\graph}{G}
\newcommand{\nodes}{V}

\newcommand{\numnodes}{N}

\newcommand{\unode}{i}

\newcommand{\edges}{E}

\newcommand{\xinstance}{\set{x}}
\newcommand{\Xvariable}{\set{X}}
\newcommand{\Yvariable}{\set{Y}}

\newcommand{\Lvariable}{\set{L}}

\newcommand{\gprob}{\mathbb{P}}

\newcommand{\linstance}{\set{z}}
\newcommand{\ldim}{d}

\newcommand{\kdim}{k}

\newcommand{\nodem}{u}
\newcommand{\noden}{v}
\newcommand{\A}{\set{A}} 
\renewcommand{\L}{\set{L}} 
\newcommand{\numlabels}{l}

\newcommand{\X}{\set{X}} 
\newcommand{\numatts}{k}
\newcommand{\Z}{\set{Z}}
\newcommand{\Xvars}{\mathcal{X}}
\newcommand{\Avars}{\mathcal{A}}
\newcommand{\Lvars}{\mathcal{L}}
 \newcommand{\restrict}[2]{#1_{#2}}
 
\newcommand{\model}{M}
\newcommand{\aparameters}{\theta} 
\newcommand{\pparameters}{\psi} 
\newcommand{\efpars}{\phi} 

\newcommand{\elbo}{\textsl{ELBO}}

\usepackage{latexsym}
\usepackage{amssymb}
\usepackage{amsmath}
\usepackage{amsthm}
\usepackage{booktabs}
\usepackage{enumitem}
\usepackage{graphicx}
\usepackage{color}

\newcommand{\BibTeX}{B\kern-.05em{\sc i\kern-.025em b}\kern-.08em\TeX}

\begin{document}


\begin{frontmatter}


\paperid{123} 




\author[A]{\fnms{Erfaneh}~\snm{Mahmoudzadeh}
\thanks{Corresponding Author. Email:ema61@sfu.ca.}
}
\author[A]{\fnms{Parmis}~\snm{Naddaf}
}
\author[A]{\fnms{Kiarash}~\snm{Zahirnia}
} 
\author[A]
{\fnms{Oliver}~\snm{Schulte}
} 

\address[A]{School of Computing Science, Simon Fraser University}

\title{Deep Generative Models for Subgraph Prediction}
\begin{abstract} 
Graph Neural Networks (GNNs) are important across different domains, such as social network analysis and recommendation systems, due to their ability to model complex relational data. This paper introduces subgraph queries as a new task for deep graph learning. Unlike traditional graph prediction tasks that focus on individual components like link prediction or node classification, subgraph queries jointly predict the components of a target subgraph based on evidence that is represented by an observed subgraph. For instance, a subgraph query can predict a set of target links and/or node labels.  To answer subgraph queries, we utilize a probabilistic deep Graph Generative Model. Specifically, we inductively train a Variational Graph Auto-Encoder (VGAE) model, augmented to represent a joint distribution over links, node features and labels. Bayesian optimization is used to tune a weighting for the relative importance of links, node features and labels in a specific domain. 
We describe a deterministic and a sampling-based inference method for estimating subgraph probabilities from the VGAE generative graph distribution, without retraining, in zero-shot fashion. For evaluation, we apply the inference methods on a range of subgraph queries on six benchmark datasets. We find that  inference from a model achieves superior predictive performance, surpassing independent prediction baselines with improvements in AUC scores ranging from 0.06 to 0.2 points, depending on the dataset. 
\end{abstract}

\end{frontmatter}

\section{Introduction: Subgraph Queries}

Due to the importance and ubiquity of graph data, deep graph learning is a growing field with many applications. Typical graph learning tasks include link prediction (predicting the existence of a target link given as evidence other links and node attributes) and node classification (predicting a node label given the node links and labels of other nodes). This paper addresses \textsl{subgraph prediction}, which is a significant novel generalization of the link prediction and node classification tasks ~\citep{hamilton2020graph}. A subgraph query aims to estimate the probability of a target subgraph (i.e., a set of links, node labels, and node features), given an observed subgraph as evidence. Since subgraph components are not independent of each other, optimal subgraph prediction is different from predicting each target link/node label independently. We introduce VGAE+, an extension of the VGAE base generative model~\cite{Kipf2016}. To answer subgraph queries, the trained VGAE+ models takes an attributed evidence graph as input and returns a  probability for the target subgraph. 



\begin{figure}[ht]
    \centering
    \includegraphics[width=\columnwidth]
    {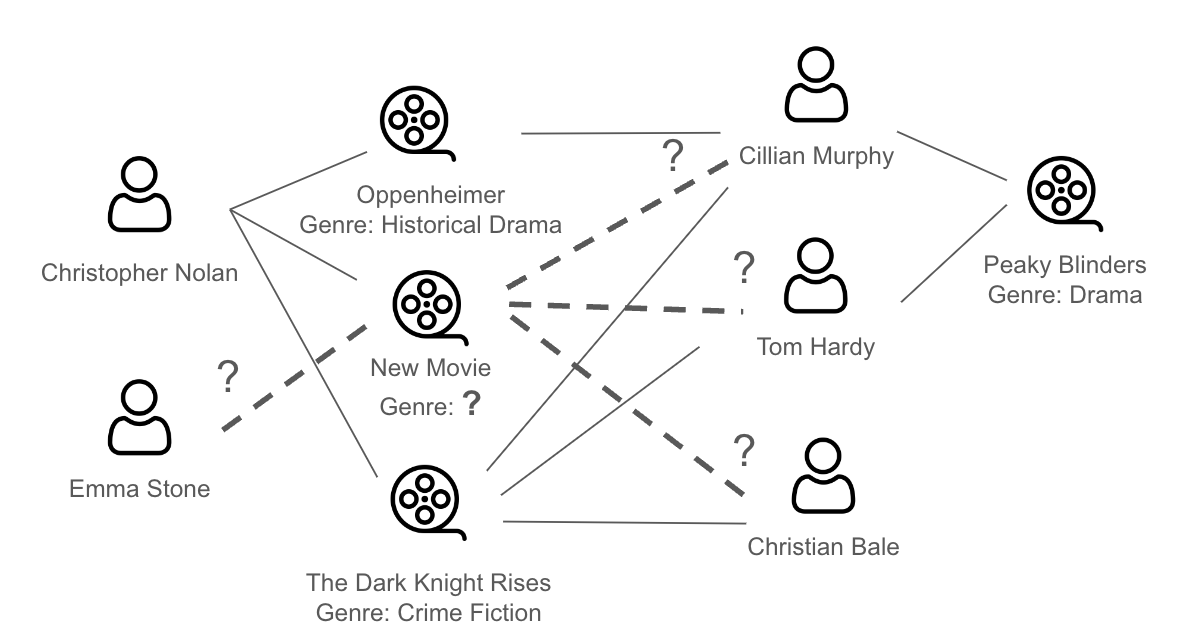}
    \caption{Example of joint link prediction and node classification: predicting potential actors for Christopher Nolan's new movie and its genre. Target links are dashed lines marked with "?". Solid links are specified as evidence.\\}
    \label{fig:example}
\end{figure}

\textsl{Examples. } We provide some examples to illustrate the power of subgraph prediction, describe use cases, and clarify the relationship to link prediction and node classification. 
\Cref{fig:example} shows a scenario with a new film directed by Christopher Nolan. Previously, he collaborated with Cillian Murphy, Christian Bale, and Tom Hardy in "The Dark Knight Rises," and with Cillian Murphy in "Oppenheimer." Suppose our evidence stipulates that  Cillian Murphy is cast in this new movie, and we want to predict which previous collaborators will join him in the movie, and which type of movie it will be.   The subgraph prediction task is to jointly predict (i) the movie's genre and (ii) which additional actors will join the cast. 

Predicting the genre of the new movie 
is an instance of {\em single node classification}, an extensively studied task~\citep{xiao2022graph, cai2018comprehensive}. Predicting whether Tom Hardy will be in the new movie (independent of other actors) is an instance of {\em single link prediction}, an equally extensively studied task ~\citep{hamilton2020graph}.  Predicting whether Tom Hardy, Cillian Murphy, and Christian Bale will all join the movie, is an instance of \textsl{joint link prediction}, a recently introduced task~\citep{naddaf2023joint} that generalizes single link prediction.  Our example query illustrates how subgraph prediction generalizes both joint link prediction and node classification.

For another example, consider fraud detection in financial networks. A node classification approach is to label a network node as  ``suspicious" or ``normal"~\citep{tang2024gadbench}.
However, recent attack types involve collusion among many network nodes controlled by the attackers~\citep{li2021}.
Detecting collusion can be achieved by a more powerful approach based on joint node classification. Moreover, attack nodes exchange messages at a substantially higher frequency than normal nodes~\citep{vitorino2023sok}.
Subgraph prediction can capture joint association patterns among node labels and links to model both which nodes are suspicious and how suspicious nodes interact with each other. 

\textbf{Motivation.} Subgraph queries are substantially more expressive than the previously studied link prediction and node classification tasks. 
They allow users to choose as prediction target a {\em set} of links and a {\em set} of node features or labels rather than assuming a fixed-size prediction target (e.g., single link, single label), instead of choosing one of them. Second, users can provide different kinds of evidence, depending on what is known, and use a single system for answering the query. In contrast, previous researchers have developed a separate customized method for different evidence types. Answering general relational queries, including subgraph queries, is a major use case for non-neural statistical-relational models, such as Markov Logic networks~\citep{Domingos2009},
which 
motivates our goal of expanding the prediction tasks to which neural models can be applied.

Recent generative AI models have shown the usefulness for many users of a single multi-task system.
For deploying graph learning, a single query answering system is important in a production environment where we do not know in advance which graph queries will be important, and users may not have the resources to build a customized machine learning solution for different query types. In this work, we are proposing VGAE+, a generative model that supports inference to return a probabilistic answer to a subgraph query whose target and evidence set can be specified by the user. A generative model, unlike a discriminative model, supports prediction with targets of varying sizes. 
\begin{figure}[ht]
\centering
\includegraphics[width=0.43\textwidth]{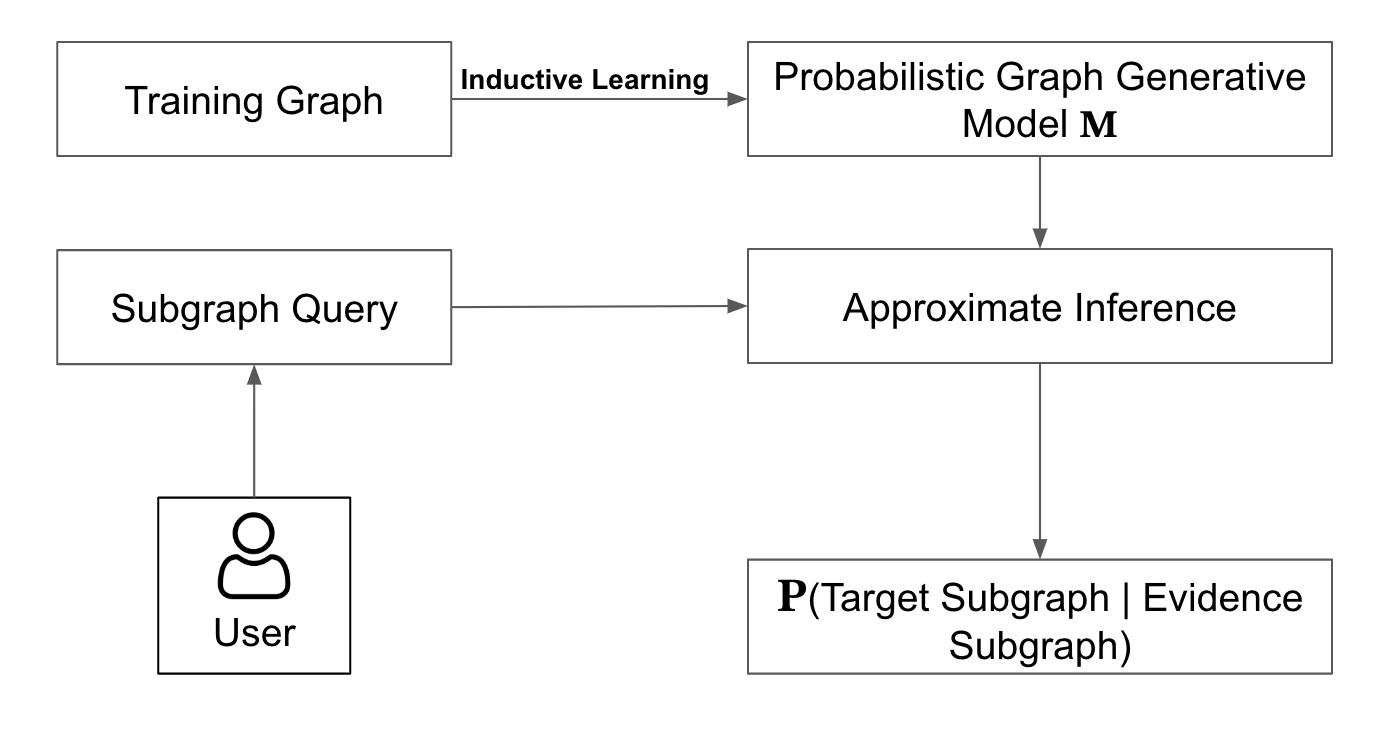}
\caption{After training a single GGM, the approximate inference methods described in this paper can answer a user query.\\} \label{fig:query-flow}
\end{figure} 

\paragraph{Approach.}  
\Cref{fig:query-flow} shows our system design. 
Our approach is a form of {\em domain-specific pre-training}: 
 We learn a probabilistic Graph Generative Model (GGM) from data in a deployment domain. The model is used to answer queries with no further learning required. The deployment domain can be large (e.g., 1M nodes), whereas queries are typically small (e.g., involving 10-100 nodes).
 Specifically, we train a generative Variational Graph Auto-Encoder (VGAE)~\citep{Kipf2016,hamilton2020graph} inductively, so that the VGAE can be applied to query targets of
 different sizes that may involve unseen nodes. 
 The generative probabilities over graphs (implicitly) define a conditional probability $P(\mathit{target\_subgraph} |\mathit{evidence\_subgraph})$ for every subgraph query. 
 The VGAE utilizes an encoder-decoder architecture with a Graph Neural Network (GNN) as an encoder, {\em augmented} to generate node labels/features as well as links. 
We describe and evaluate approximate inference methods for estimating conditional subgraph probabilities from a trained VGAE. 

\paragraph{Evaluation.} 
We evaluate a variety of query types on six benchmark homogeneous graphs,  depending on whether 1) the target subgraph involves a single neighbor or multiple neighbors of a target node, and 2) the evidence comprises only links among training nodes (transductive), or also links that connect training and test nodes (inductive).
 Our main baseline is to estimate link probabilities and node labels independently using SOTA node classification/link prediction methods. For subgraph queries, we find that the VGAE model achieves superior or competitive performance on all datasets and settings, especially inductive inference. 
For node classification/link prediction problems, we find that inference from the augmented VGAE model is competitive with custom methods.

\paragraph{Contributions.} Our main contributions are as follows. 

\begin{itemize}
    \item Introducing the task of conditional subgraph prediction. 
    \item Approximate inference methods, including sampling, for answering subgraph queries from a single trained GGM.
    \item Introducing VGAE+, which is an augmented VGAE for jointly modeling links and node features/labels. VGAE+ is trained with a new objective that utilizes Bayesian optimization to weight link prediction and node classification. 
\end{itemize}

{\em Paper Organization.} We review related work, then define subgraph queries. 
We describe inductive training of a VGAE based on the Evidence Lower BOund (\elbo)  likelihood approximation and inference methods for answering subgraph queries from the trained VGAE. An extensive empirical evaluation compares subgraph prediction from a VGAE model to 6 baselines on 6 benchmark datasets.
 
\section{Related Works} \label{sec:related}

{\em Link Prediction.} 
For deep graph models, the most common setting is {\em single link completion}: predict a single target link given a set of known links, which typically include a large set of training links. Other recent variants include the following single link tasks. (1) Condition on the attributes of the two target nodes only~\citep{hao2020inductive}. (2) Condition on links among other test nodes only~\citep{teru2020inductive}. (3) Condition on links among training and test nodes~\citep{zhang2019inductive}. These single-link tasks are generalized by the recently introduced joint link prediction task~\citep{naddaf2023joint}. In joint link prediction, the target is a set of links to be predicted jointly, given another set of links and node features as evidence. Joint link prediction is the closest predecessor to our work in that it aims to answer a large class of graph queries from a single model. However, joint link prediction does not provide the capability of predicting node labels or features, which is important in many applications. Subgraph prediction requires training a generative model to support the dual task of joint link prediction and joint node classification, which we address with Bayesian optimization in this paper. 
%

{\em Node Classification.} 
Node classification is one of the most common tasks in graph analysis. The  goal is to predict a class for each unlabelled node in the graph based on available graph evidence~\citep{KAZIENKO2012199}, which usually includes all links and node features. Graph neural networks (GNNs) achieve state-of-the-art performance~\citep{xiao2022graph}. We include a recent method~\cite{hassani2020contrastive} in our baseline comparisons. The common semi-supervised node classification task is similar to a subgraph query in that it requires a joint prediction of node labels. It is different in that a subgraph query requires predicting links as well.

{\em Two-Task GNNs vs. Inference from a Model.} There is a recent trend towards two-task GNNs that support both node classification and link prediction~\cite{WU2022102902}. We include the recent two-task method GiGaMAE among our baselines \cite{shi2023gigamae}. While two-task GNNs are designed to support both node classification and link prediction, they require separate training for each. In contrast, we train a single model for the large set of diverse tasks that can be represented as subgraph queries. To our knowledge, inference from a generative model to estimate subgraph probabilities is a new  application of GGMs. The only other paper that addresses inference from a deep GGM is~\cite{naddaf2023joint}, which covers link prediction only. We develop a novel augmented variant of the VGAE to generate node labels and features as well as links. The Graph Variational Auto Encoder (GVAE)~\cite{simonovsky2018graphvae} also generates node labels and features, but it cannot be applied to a single training graph. Also the GVAE does not utilize Bayesian optimization to balance the relative importance of modeling links vs. node labels/features. 

As we mentioned in the introduction, statistical-relational models also support subgraph queries, but they are based on very different assumptions and model classes (e.g. exponential 
graph models) from deep GGMs, so we leave a direct comparison for future work. 

{\em Graph Queries.}
 Powerful (non-probabilistic) graph query languages have been developed, such as Cypher and SPARQL~\citep{francis2018cypher,prudhommeaux2008sparql,galkin2022inductive}. Such graph queries typically return a set of nodes or links that satisfy a  complex condition. 
 In contrast, subgraph queries return a probability for a target subgraph. 

{\em Inductive Graph Training.} Inductive graph learning has been a major topic in recent research~\citep{hamilton2017inductive,zeng2019graphsaint,rossi2018deep}.
%
%
To support link prediction for nodes not seen during training, our query answering approach can be used with any inductive encoder-decoder graph neural network architecture, 
trained with the variational \elbo~ objective described below. 
To our knowledge, this is the first use of the variational \elbo~ objective to support inductive subgraph prediction.




\emph{Graph Generative Models.} Inference from a model requires the following properties of a GGM. (1) Supports the computation/estimation of explicit graph probabilities. (2) Admits a conditional variant.
(3) Applies to graphs of different sizes. As we show in this paper, VGAEs meet these requirements, so we base our experiments on them. Specifically, we employ the most recent VGAE designed for link prediction~\citep{naddaf2023joint}. Another advantage of the VGAE is that it is designed for a single large dataset, as are most methods for link prediction and node classification, so we can employ the same benchmark datasets. In contrast, other deep GGMs are usually trained on datasets with many disjoint graphs~\citep{faez2021deep} (e.g., molecules).
We believe that developing deep GGMs so that they support answering subgraph queries is a fruitful new direction for GNNs.

{\em Subgraph Classification.}~\citeauthor{alsentzer2020subgraph} \citep{alsentzer2020subgraph} identify subgraph prediction as an important task.
Like our approach, their query targets and evidence are subgraphs. However, their system addresses only {\em classifying} the target subgraph with a single label. In contrast, we assign a subgraph probability.  

{\em Structure-Conditioned Graph Generation} is a recently studied variant where the user specifies an induced subgraph (the evidence) and a conditional graph generation process completes it~\citep[Sec.D.1]{vignac2022digress},~\citep{faez2022scgg}.
Our definition of joint link prediction allows that neither evidence nor target links define a complete graph, and that they may share nodes, which raises special challenges (see~\Cref{sec:inference}). In addition, our aim is to output an explicit probability over a joint link assignment, not output a graph. The fact that this recent work has studied a related task beyond single component prediction supports our motivation for more complex queries.





\section{Subgraph Queries} 
\label{sec:subgraph_query}

In this section we formally define our problem, answering probabilistic subgraph queries (SQs). SQs specify a set of links and attributes for a set of query nodes. The specification can involve a language as complex as first-order logic~\citep{Domingos2009}. 
In this paper we introduce a relatively simple SQ syntax---essentially, conjunctions of links, and/or node attributes and labels---that suffices to express the prediction tasks that have been previously studied in graph learning (~\Cref{sec:related}). 

\paragraph{Data Format.}

An attributed labelled graph is a pair $\graph=(\nodes,\edges)$ comprising a finite set of nodes $\nodes$ with features and labels, and links $\edges$, which may be positive (present) or negative (absent). Each node  is assigned a $\numatts$-dimensional attribute $\xinstance_{\unode}$ with 
$\numatts>0$ and a node label $\set{l}_{i}$. A graph with $\numnodes$ nodes can be represented by the following objects.

\begin{itemize}
    \item An $\numnodes \times \numnodes$ adjacency matrix $\A$ with $\{0,1\}$ 
Boolean entries.
\item An $\numnodes \times \numatts$ node feature matrix $\X$ with $\{0,1\}$ 
Boolean entries.
\item An $\numnodes \times \numlabels$ node label matrix $\L$ with a one-hot encoding of $\numlabels$ discrete labels for each node.  
\end{itemize}

A target subgraph of graph $\graph=(\nodes,\edges)$ is a graph $\target{\graph}=(\target{\nodes},\target{\edges})$ 
where $\target{\nodes} \subseteq \nodes$ and $\target{\edges} \subseteq \edges.$ Similarly, an evidence subgraph of graph $\graph$ is a graph $\evidence{\graph}=(\evidence{\nodes},\evidence{\edges})$ 
where $\evidence{\edges} \subseteq \edges$ and $\evidence{\edges} \cap \target{\edges} = \emptyset.$
We refer to the nodes $\evidence{\nodes}$ that appear in the evidence as \defterm{evidence nodes}, to the nodes $\target{\nodes}$ that appear in the target as \defterm{target nodes}, and to their union as \defterm{query nodes} (i.e., $\nodes = \evidence{\nodes}\cup \target{\nodes}$).

\paragraph{Definition of Subgraph Queries.} \label{sec:graph-queries}

A relational random variable corresponds to either a node attribute or an adjacency. Let $\Avars = \{\A[\nodem,\noden]: \nodem \in \nodes, \noden \in \nodes\}$ be the set of \defterm{link variables},  and $\Lvars = \{\Lvariable[\nodem]: \nodem \in \nodes\}$ be the set of \defterm{node label variables}, and $\Xvars = \{\Xvariable[\nodem]: \nodem \in \nodes\}$ be the set of \defterm{feature variables}. 
%




A \defterm{subgraph query} $\gprob(\mathit{target}|\mathit{evidence})$ is of the form 

\begin{equation}
\label{eq:sgq}
\resizebox{\columnwidth}{!}{$
    \gprob(\target{\Avars}=\target{\set{a}}, \target{\Lvars}=\target{\set{l}},\target{\Xvars}= \target{\set{x}}|\evidence{\Avars} = \evidence{\set{a}}, \evidence{\Lvars}=\evidence{\set{l}}, \evidence{\Xvars} = \evidence{\set{x}}) \nonumber$}
\end{equation}

where 

\begin{itemize}
    \item $\target{\Avars} = \{\A[\nodem_{i},\noden_{i}]: i = 1,\ldots,|\target{\Avars}|, \A[\nodem_{i},\noden_{i}] \in \target{\edges} \}$ is the list of binary target links, each assigned a value $a_i \in \{0,1\}$. Similarly, $\evidence{\Avars} = \{\A[\nodem_{i},\noden_{i}]: i = 1,\ldots,|\evidence{\Avars}|, \A[\nodem_{i},\noden_{i}] \in \evidence{\edges}\}$ is the list of binary evidence links. 
    \item $\target{\Lvars} = \{\Lvariable[\nodem_{i}]: i = 1,\ldots,|\target{\Lvars}|, \nodem_{i} \in \target{\nodes} \}$ is the list of target node labels, each assigned a one-hot encoding of the $\numlabels$ class labels $l_i \in \{0,1\}^\numlabels$. Similarly, $\evidence{\Lvars} = \{\Lvariable[\nodem_{i}]: i = 1,\ldots,|\evidence{\Lvars}|, \nodem_{i} \in \evidence{\nodes}\}$ is the list of evidence node labels. 
   \item $\target{\Xvars} = \{\Xvariable[\nodem_{i}]: i = 1,\ldots,|\target{\Xvars}|, \nodem_{i} \in \target{\nodes}\}$ is the list of target node features, each assigned a feature vector $\set{x}_i \in \mathbb{R}^{1 \times \numatts}$. Similarly, $\evidence{\Xvars} = \{\Xvariable[\nodem_{i}]: i = 1,\ldots,|\evidence{\Xvars}|, \nodem_{i} \in \evidence{\nodes}\}$ is the list of evidence node features.
\end{itemize}

The indices $i$ index elements in a query not nodes in a graph. 
Given a partition of nodes into observed 
training nodes and 
unobserved 
test nodes, a query is \defterm{inductive} if a test node appears in the target nodes. 

Note that the target and evidence specifications can be and typically are {\em partial} in that some graph components are unspecified. For example, for two evidence nodes $\nodem$ and $\noden$, the evidence may specify their features, but not whether there exists a link between them or not. 
Our definition 
treats a node feature $\set{x}[\nodem]$ as a group, in that either all or no features of node $\nodem$ are specified in a query. This restriction is not essential; we use it mainly to simplify notation. 

\begin{figure}[ht]
\centering
    \begin{subfigure}[b]{0.23\textwidth}
        \centering
        \includegraphics[width=\textwidth]{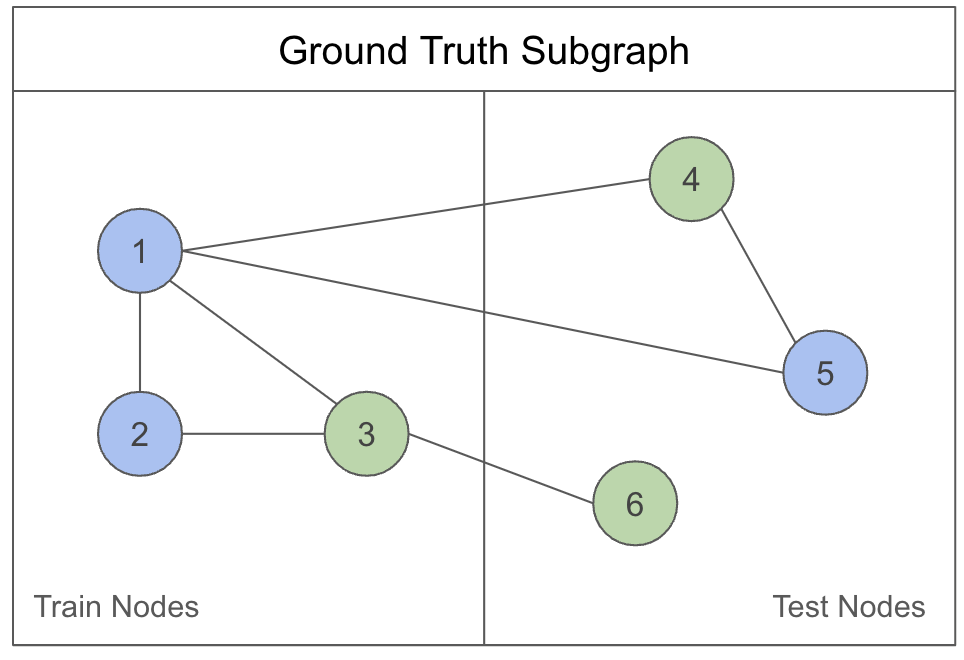}
        \label{fig:input_graph}
    \end{subfigure}
    \begin{subfigure}[b]{0.23\textwidth}
        \centering
        \includegraphics[width=\textwidth]{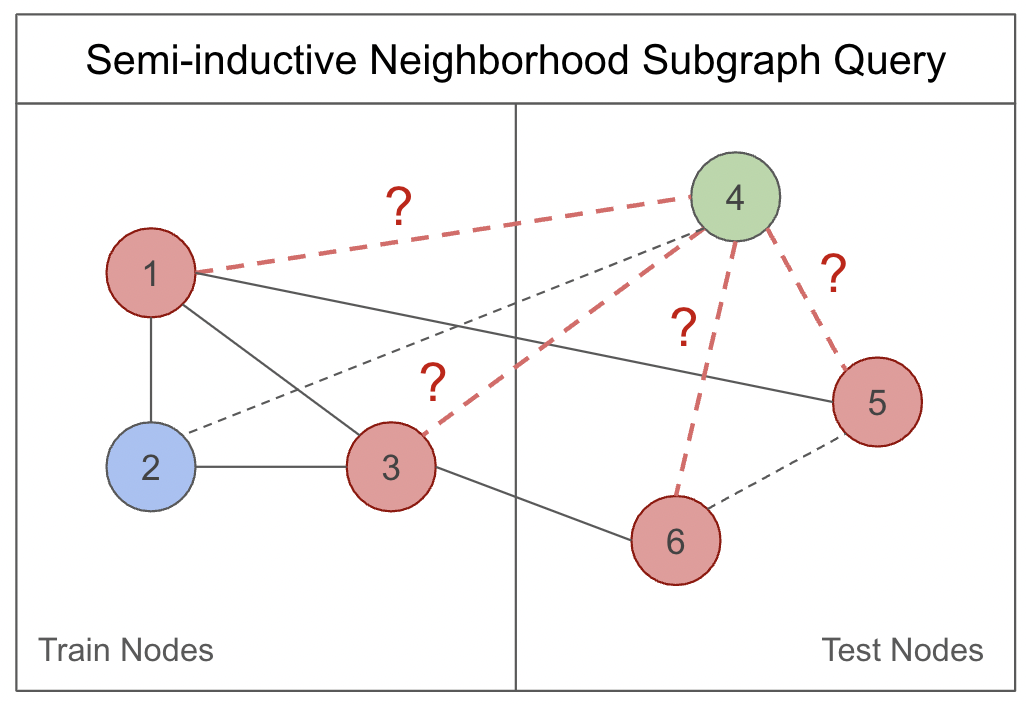}
        \label{fig:query_example}
    \end{subfigure}
    \caption{\textbf{Left}: Input graph with partition of nodes. Node labels are green and blue. \textbf{Right}: Neighborhood query: Target labels and links are colored red. Black dashed links are unspecified links in evidence.}
    \label{fig:split_example}
\end{figure}

\paragraph{Example}
Figure~\ref{fig:split_example}(Right) shows an example of the most complex SQ type we consider in this research. 

\begin{itemize}
    \item The query nodes are 1--6.
    \item The {\em query target subgraph} comprises:
    \begin{itemize}
    \item two positive target links defined by the pairs $(1,4),(5,4)$.
    \item two negative target links defined by the pairs $(3,4),(6,4)$. 
        \item the labels of nodes $1,3,5,6$, 
    \end{itemize}
    \item The {\em query evidence subgraph comprises}:
    \begin{itemize}
       \item five positive evidence links defined by the pairs $(1,2),(1,3),(2,3),(3,6),(1,5)$. 
    \item four negative evidence links defined by the pairs $(3,5),(2,6),(2,5),(1,6)$. 
    \item The features of all query nodes (not shown). 
    \end{itemize}
    \item The presence/absence of links $(2,4)$ and $(6,4)$ is unspecified. 
\end{itemize}

The perfect query answer is to assign probability 1 to the ground truth subgraph shown on the Figure~\ref{fig:split_example}(Left). The query is inductive because nodes 4,5,6 are not observed during model training.

\vskip 10mm

\section{Variational Graph Auto-Encoder
Training} \label{sec:generative}
\begin{figure}[ht]
\centering
\includegraphics[width=0.48\textwidth]{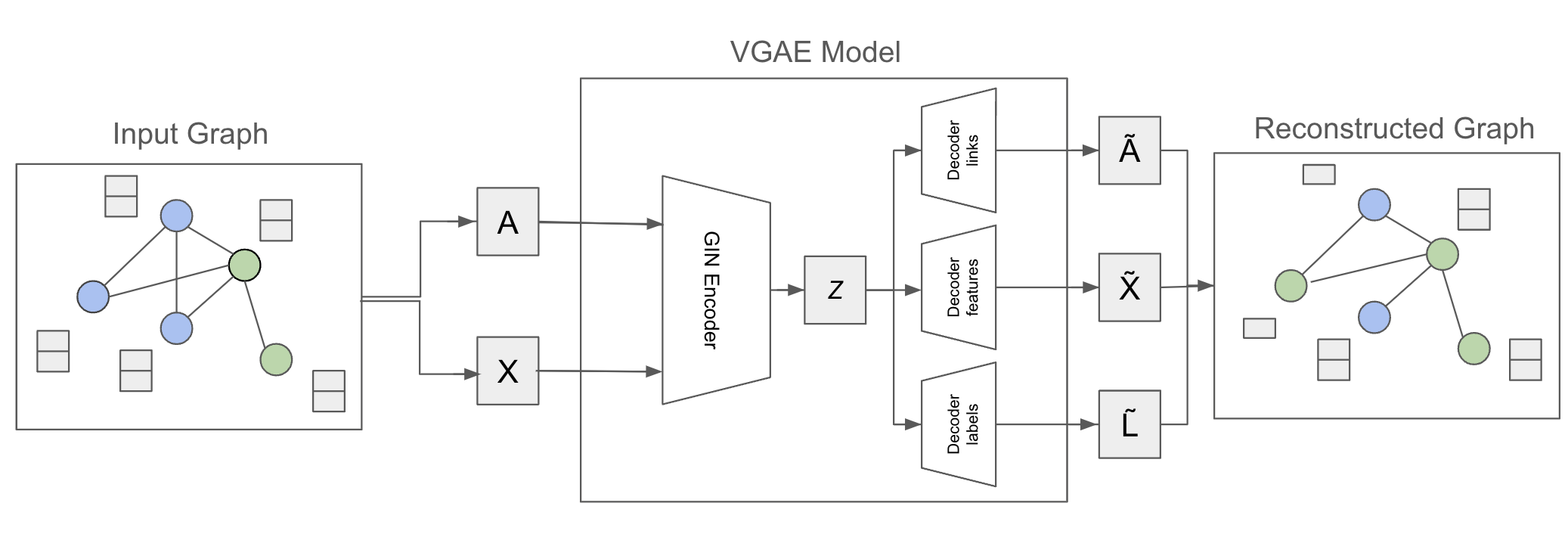}
\caption{Encoder-Decoder Training Architecture}
\label{fig:model}
\end{figure}

We describe our generative model, a VGAE augmented with feature and label reconstruction, with training and implementation details.~\Cref{fig:model} shows the VGAE training architecture.

\subsection{Augmented VGAE Generative Model}


Let $\Z$ be an $\numnodes \times \ldim$ matrix that represents latent node embeddings. 
In the VGAE model, links are generated independently given node embeddings. Following the GraphVAE approach~\citep{simonovsky2018graphvae}, we generate node classes and node features independently as well given node embeddings. We thus utilize three decoder models (see~\Cref{fig:model}):

\begin{equation}
\begin{split}
&p_{\aparameters}(\A|\Z) = \prod_{\nodem,\noden}  p_{\aparameters}(\A[\nodem,\noden]|\linstance[\nodem],\linstance[\noden])
\\
&p_{\pparameters}(\X|\Z) = \prod_{\nodem}  p_{\pparameters}(\Xvariable[\nodem]|\linstance[\nodem])
\\
&p_{\efpars}(\Lvariable|\Z) = \prod_{\nodem}  p_{\efpars}(\Lvariable[\nodem]|\linstance[\nodem])
\label{eq:mixture}
\end{split}
\end{equation}

where $p_{\aparameters}: \mathbb{R}^{\ldim} \times \mathbb{R}^{\ldim}  \rightarrow [0,1]$ is a trainable \defterm{link decoder}, and $p_{\pparameters}$ resp. $p_{\efpars}$ denotes a trainable \defterm{feature decoder} resp. \defterm{label decoder}.

The graph \defterm{encoder} $q_{\efpars}(\Z|{\Xvariable},\A)$ is implemented by a GNN that takes as input an attribute graph and returns latent node embeddings. For compatibility with baseline methods, the encoder does not receive node labels as input, but adding them is straightforward.  

A VGAE+ 
is trained using the variational \elbo~ objective~\citep{Kipf2016,hamilton2020graph}: 

\begin{equation}
\begin{split}
&\loss(\aparameters,\efpars) =  - E_{\Z \sim q_{\efpars}(\Z|\Xvariable,\A)}[
 \alpha \times \ln{ p_{\aparameters}(\A|\Z})
\\
&+\beta \times \ln{ p_{\pparameters}(\X|\Z})
+\gamma \times \ln{ p_{\efpars}(\L|\Z})\big] 
\\
&+KL\big(q_{\efpars}(\Z|\Xvariable,\A)||p(\Z)\big) 
\end{split}
\label{eq:gen-loss}
\end{equation}

where $KL(.||.)$ is the Kullback-Leibler divergence  between two probability distributions. Here $\alpha$, $\beta$ and $\gamma$ are hyperparameters that weight the importance of different reconstruction tasks. These hyperparameters are found by Bayesian optimization~\cite{bib:bayesopt}. The Optimizer tunes these hyperparameters so that they minimize the model reconstruction loss on the validation graph. For details please see the Appendix.


\paragraph{Inductive Training} \label{sec:inductive}
In order to answer inductive queries involving unseen nodes, we train the VGAE+ without node IDs. We randomly partition the nodes in the input graph into 1) 70\% training nodes $\train{\nodes}$, 
2) 20\% inductive test nodes 
$\test{\nodes}$ 
and 3) 10\% validation nodes. We train a VGAE+ model on the input subgraph induced by $\train{\nodes}$.

\subsection{Implementation}

\paragraph{Encoder} The graph \defterm{encoder} $q_{\efpars}(\Z|\X,\A)$ is implemented by a GNN that takes an attributed labelled graph as input and returns node embeddings $\Z$. The node embeddings are independent, and for each node, represent a conditional Gaussian distribution, such that 

\begin{equation} \label{eq:evidence}
    q(\linstance[\nodem]|\X,\A)\sim N(\mu[\nodem],\sigma[\nodem])
\end{equation}
  
 with mean and covariance $\mu[\nodem],\sigma[\nodem]$ for 
 node $\nodem$. 

\paragraph{Link Decoder}



As a strong link decoder, we utilize a Stochastic Block Model (SBM), which is defined as:

\begin{equation} \label{eq:link-decoder}
p_{\aparameters}(\A[\nodem,\noden]|\linstance[\nodem],\linstance[\noden]) = \linstance[\nodem]^\top\Lambda \linstance[\noden]
\end{equation}

where $\Lambda\in R^{d\times d}$ is the trainable $d$-block matrix in the SBM.


\paragraph{Feature Decoder}
Nodes features are reconstructed independently, given node embeddings. Our benchmark datasets comprise discrete features, so our {\em feature decoder}  
is of the form $p_{\pparameters}: \mathbb{R}^{\ldim}  \rightarrow \{0,1\}^{\kdim}$ 


\begin{equation}
p_{\pparameters}(\Xvariable|\Z) 
= \prod_{u=1}^{\numnodes}  \prod_{d=1}^{k} p_{\pparameters}(\Xvariable_{d}[\nodem]|\linstance[\nodem]) = \prod_{\nodem=1}^{\numnodes}  \prod_{d=1}^{k} \sigma(\tilde{x}_{d}[\nodem]),
\label{eq:mixture_X}
\end{equation}

where $\pparameters$ are the parameters of a fully connected neural net feature decoder that maps a node embedding $\linstance[\nodem]$ to a $k$-dimensional node feature reconstruction $\tilde{x}[\nodem]$.


\paragraph{Node Classifier}
The {\em node classifier} $p_{\efpars}: \mathbb{R}^{\ldim}  \rightarrow \{0,1\}^{\numlabels}$, classifies the  node labels independently under a one-hot encoding, given the same node embeddings:
\begin{equation}
p_{\efpars}(\Lvariable|\Z) = \prod_{\nodem=1}^{\numnodes}  p_{\efpars}(\Lvariable[\nodem]|\linstance[\nodem])=\prod_{\nodem=1}^{\numnodes}  \mathit{softmax}(\tilde{l}[\nodem]) 
\label{eq:mixture_L}
\end{equation}
where $\efpars$ are the parameters of a fully connected 
node classifier that maps a node embedding $\linstance[\nodem]$ to a $\numlabels$-dimensional vector $\tilde{l}[\nodem]$. 


 



\section{Subgraph Inference from a VGAE Model} 

Consider a subgraph query~\Cref{eq:sgq} with $n$ query nodes 

$\gprob(\target{\Avars}=\target{\set{a}}, \target{\Lvars}=\target{\set{l}},\target{\Xvars}= \target{\set{x}}|\ev = \evinstance)$ where $\ev$ is a list of evidence variables. 
We define two subgraph inference models, deterministic and Monte Carlo, for the VGAE+ generative model. Essentially, the deterministic method uses node embeddings computed deterministically from the evidence, whereas the MC method samples node embeddings conditional on the evidence. 

\subsection{Inference Models} \label{sec:inference}

For a fixed set of $n$ query node embeddings $\linstance$, the target probability is given by

\begin{equation}
\begin{split} \label{eq:target-z}
&\gprob_{\set{\eta}}(\target{\Avars}=\target{\set{a}}, \target{\Lvars}=\target{\set{l}},\target{\Xvars}= \target{\set{x}}|\linstance) 
\end{split} 
\end{equation}

and can be computed by multiplying the independent decoder probabilities~\cref{eq:link-decoder,eq:mixture_X,eq:mixture_L} parametrized by $\set{\eta}$. 
The \defterm{posterior distribution} over the $n$ node embeddings is a conditional Gaussian distribution, such that $p(\linstance[\nodem]|\ev=\evinstance)\sim N(\mu[\nodem],\sigma[\nodem])$  
 with mean and covariance $\mu[\nodem],\sigma[\nodem]$ for 
 node $\nodem$. ~\Cref{eq:prior} below shows how to approximate the posterior using the GNN encoder. 
Let $\set{\mu}_{\ev}$ be the mean node embeddings from the posterior distribution.  \textbf{Deterministic inference}  utilizes the posterior mean embeddings: 

\begin{align} \label{eq:det}
     \gprob(\target{\Avars}, \target{\Lvars},\target{\Xvars}|\ev = \evinstance) \approx \gprob(\target{\Avars}, \target{\Lvars},\target{\Xvars}|\set{\mu}_{\ev}).
\end{align}

\paragraph{Monte Carlo (MC) inference} samples ${S}$ node embeddings from the posterior distribution, and averages the subgraph probabilities:

\begin{align} \label{eq:mc}
     \gprob(\target{\Avars}, \target{\Lvars},\target{\Xvars}|\ev = \evinstance) \approx \frac{1}{S} \sum_{s=1}^{S}  \gprob(\target{\Avars}, \target{\Lvars},\target{\Xvars}|\linstance^{s})
\end{align}

where $\Z^{s}\sim \prod_{\nodem = 1}^{n} p(\linstance[\nodem]|\ev=\evinstance)$. 

\subsection{Computing the Posterior Distribution} 
Training a VGAE+ model $\model$ provides decoder models $p_{\set{\eta}}$ and an encoder $q_{\efpars}$.
%
The challenge is that the encoder assumes as input an adjacency matrix, or equivalently, a complete subgraph induced by the query nodes. We bridge the gap between a partially specified subgraph and a complete induced subgraph by imputing graph components missing from the evidence with 0 as a default value. For missing links, using the 0 default is the approach taken in previous work~\citep{naddaf2023joint,Kipf2016}. As discussed by~\cite{naddaf2023joint}, 0 default is appropriate for links: First, a message-passing encoder treats 0 links as uninformative, which is appropriate for for unspecified links. Second, because of graph sparsity, the mode of the true link posterior given the evidence is close to 0. For node features and labels, the 0 default is also appropriate given our one-hot encoding, since the VGAE+ encoder is trained not to propagate information from 0-valued features/labels. 

Formally, to obtain \textbf{evidence embeddings}, we approximate the posterior distribution using the trained encoder $q_{\efpars}$:

\begin{align} \label{eq:prior}
  p(\Z|\ev = \evinstance) \approx 
  q_{\efpars}(\Z|\A^{\ev,\set{0}},\Xvariable^{\ev,\set{0}}) 
\end{align}

where the {\em evidence matrices} $\A^{\ev,\set{0}}$ and $\Xvariable^{\ev,\set{0}}$ are defined as follows: 1) The dimension of $\A^{\ev,\set{0}}$ is $n \times n$. 
If $\evidence{\Avars}$ specifies a link assignment $\Avars[\nodem,\noden] = e_{i}$, then $\A^{\ev,\set{0}}[\nodem,\noden] := e_{i}$; otherwise $\A^{\ev,\set{0}}[\nodem,\noden] := 0$. 2) The dimension of $\Xvariable^{\ev,\set{0}}$ is $n \times \numatts$. If $\evidence{\Xvars}$ specifies a feature vector $\Xvars[\nodem] = \set{x}_{i}$, then $\Xvariable^{\ev,\set{0}}[\nodem] := \set{x}_{i}$; otherwise $\Xvariable^{\ev,\set{0}}[\nodem] := \set{0}$. Similarly we can assign a default value of 0 to unspecified node labels, but our experiments do not utilize queries that include node labels as evidence. Note that while the number of training nodes may be very large, the number of query nodes is typically small (on the order of 10-100). Applying the same encoder to subgraphs of different sizes is possible because we train the VGAE+ model inductively. (For details please see the Appendix.)

To illustrate, consider~\Cref{fig:split_example}. Since the link between nodes 1 and 2 is specified to exist, the evidence adjacency matrix assigns $\A^{\ev,\set{0}}[1,2] := 1$. The link between nodes 3 and 5 is specified not to exist, so the evidence matrix assigns $\A^{\ev,\set{0}}[3,5] := 0$. 
Since the link between nodes 5 and 6 is unspecified, the evidence matrix assigns $\A^{\ev,\set{0}}[5,6] := 0$. 
If the feature vector for node 2 is unspecified, the evidence matrix assigns the zero feature vector: $\Xvariable^{\ev,\set{0}}[2] := \set{0}$.

\section{Evaluation}

We detail our methodology then discuss our empirical results. Our GitHub repository provides a  PyTorch
implementation and datasets.\footnote{\url{https://github.com/erfmah/Answering_Graph_Queries}}

\subsection{Experimental Design.}
We describe our benchmark datasets, the design of the test queries, and how evaluation metrics are computed. 

\paragraph{Datasets}
We utilize datasets from previous studies of GGMs ~\citep{Kipf2016,yun2019GTN,hao2020inductive}.
Cora, ACM, and CiteSeer are citation networks, IMDb is a movies dataset, and Photo and Computers are co-occurrence networks based on Amazon data. The appendix presents dataset statistics.

\paragraph{Data Preprocessing} 
\label{sec:data-preprocessing}
Following previous link prediction studies~\citep{Kipf2016}, we add self-loops and make all links undirected (i.e., if the training data contains an adjacency, $\noden\rightarrow \nodem$, it also contains $\nodem\rightarrow \noden$.)  Cora, CiteSeer, Photo, and Computers are homogeneous datasets, whereas ACM and IMDb are heterogeneous datasets. Since our comparison methods use homogeneous GNNs, we homogenize different non-hierarchical edge types, such that every edge in the adjacency matrix is represented by 0 for no link and 1 for link existence. 

\paragraph{Test Query Design.}
\label{sec:test-queries}
We explain next our method for generating test subgraph queries of different types. The appendix contains visual examples of the test queries.
In all our test queries, the evidence $\ev$ does not contain node labels, but specifies 1) the node feature vector for each query node, 2) the non-target links from the input graph. To define query target nodes, we randomly select a set of 100 test nodes as target nodes from the test nodes $\test{\nodes}$ (cf.~\Cref{sec:inductive}).  







\paragraph{Single Neighbor Query} For each target node $\nodem$, we randomly select two test links, one positive pair $(\nodem,\noden_+)$ from the neighborhood of $\nodem$, one negative pair $(\nodem,\noden_-)$ from outside the neighborhood.  The resulting query is of the form $P(\Lvariable[\nodem],\A[\nodem,\noden_{+}], \A[\nodem,\noden_{-}]|\ev)$: Which of the two links is true and what is the label of the target node? 
 In inductive learning, the paired nodes $\noden_+, \noden_-$ are both test nodes. In semi-inductive learning, each paired node $\noden_+, \noden_-$ may be either a training node or a test node.



\paragraph{Neighborhood Query} For each target node $\nodem$ with at least one neighbor, let $\noden^{+}_{1},\ldots,\noden^{+}_{\mathit{deg}(\nodem)}$ enumerate the nodes in the neighborhood of $\nodem$. We randomly select negative test nodes $\noden^{-}_{1},\ldots,\noden^{-}_{\mathit{deg}(\nodem)}$. The resulting query is of the form 
$\gprob(\{\Lvariable[\noden^{+}_{i}],\A[\nodem,\noden^{+}_{i}],\Lvariable[\noden^{-}_{i}],\A[\nodem,\noden^{-}_{i}]:i=1,\ldots,\mathit{deg}(u)\}|\ev):$ 
which nodes are neighbors and what are their labels? (cf.~\Cref{fig:example}.) In semi-inductive learning, the paired nodes $\noden^{-}_{1}$ are training nodes. In inductive learning, they can be either test nodes or training nodes. 



\textbf{Evaluation Metrics.}
For single-neighbor queries, we compute the mean of each metric across all test queries. For neighborhood queries, the metric is calculated separately for each query, over all the components of the query (e.g., all the neighbors to be predicted). We then report the mean of each metric across all neighborhood queries. We separately score link prediction and node classification using ROC-AUC.
ROC-AUC is a good metric for evaluating both link prediction and node classification because it provides a comprehensive and threshold-independent measure of model performance, making it more robust to class imbalance than the accuracy metric. 
Separate scoring is supported by the VGAE+ inference model~\eqref{eq:target-z}, but in fact favors the single-component prediction baselines over our joint prediction model. The reason for independent scoring is that our baseline methods are not designed for joint prediction, so we use them only to obtain scores for individual components. The joint predictive performance for a subgraph query is measured by the average AUC score over predicted links and node labels. Other metrics (Hit-Rate@20\% and F1-macro) are in the appendix.


\subsection{Query Answering Comparison Methods.}

\paragraph{Our VGAE+ approach}  We built on the VGAE implementation for joint link prediction ~\citep{naddaf2023joint}, following the authors' recommendation of using  MC inference for link prediction and a graph isomorphism network (GIN) encoder~\citep{xu2018powerful}
with 2 layers and embedding dimension 128. We extended their VGAE architecture to include the reconstruction of node features and labels, as described in~\Cref{sec:generative}. 
Node labels are predicted using the node classifier~\cref{eq:mixture_L}, trained end-to-end. 
We give results for deterministic prediction \cref{eq:det} and MC prediction \cref{eq:mc} with 30 samples.




We select baselines that are well established and represent a variety of approaches to node classification or single link prediction. Node classification methods perform joint label prediction (known as semi-supervised node classification~\citep{xiao2022graph}) like our VGAE+ model. We apply single-link prediction methods to a set of target links separately to obtain a probability for each target link (cf.~\cite{naddaf2023joint}).
We organize methods according to whether they were originally evaluated on both link prediction and node classification, or just one or the other. 

\paragraph{Two-Task Methods} train two separate link prediction and node classification models using the same GNN architecture. \\
\textbf{Generalizable Graph Masked AutoEncoder (GiGaMAE)}  ~\citep{shi2023gigamae} 
introduces a novel graph encoder based on aligning different node embeddings that respectively encode structural and feature information. We use the authors' code to obtain node label and link prediction probabilities.
We train GiGaMAE inductively for inductive queries.


\textbf{GraphSage} ~\citep{hamilton2017inductive} is an inductive model. We have used the DGL implementation~\cite{wang2019dgl}, with the supervised training mode, for node classification, where the decoder sees the node labels. 
For link prediction, we train an SBM link decoder end-to-end using their unsupervised training. 
\textbf{Graph Attention Networks (GAT) }
are a kind of GNN that computes different weights for different nodes in a neighborhood~\cite{velivckovic2017graph}.
We used the DGL implementation (GAT)~\cite{wang2019dgl}, in supervised training mode, for node classification, where the decoder is exposed to the node labels. 
The GAT embedding system was trained with the same micro-F$_{1}$ score as in the original paper~\cite{velivckovic2017graph}.
For link prediction, we train an SBM link decoder end-to-end using unsupervised training.

\paragraph{Link prediction baselines} include the following. \textbf{SEAL} is a well-known method for transductive link prediction~\citep{zhang2018link}. We train SEAL inductively by omitting node ids (cf.~\citep{teru2020inductive,naddaf2023joint}).
The basic idea of SEAL is to embed the subgraph around a target link. 
%
\textbf{DEAL} was designed for both transductive and inductive cold-start link prediction tasks~\citep{hao2020inductive}. The distinguishing feature of the DEAL approach is that at test time, it uses only the node attributes for link prediction. This makes it a strong baseline for independent prediction since it is not sensitive to the presence of evidence links. 

\paragraph{Node classification baselines} utilize the following encoders. \textbf{MVGRL} is an inductive self-supervised approach for learning representations of nodes and graphs by contrasting different structural views of graphs~\citep{hassani2020contrastive}. We use the authors' code to train a node classification model. \textbf{DeepWalk} is a transductive graph embedding method that learns node representations by treating random walks on the graph as sentences and applying word embedding techniques~\citep{perozzi2014deepwalk}. It captures structural information of the graph by mapping nodes to low-dimensional vectors in a continuous space. The reported results are based on the DGL implementation~\cite{wang2019dgl}. Since DeepWalk cannot be adapted for inductive queries, we trained it transductively on the complete input graph.
Node classification is performed by a logistic regression model~\cite{perozzi2014deepwalk}.

\subsection{Experimental Results}

We discuss our results for different types of test queries, starting with the most complex subgraph queries. It is important to note that {\em while VGAE+ addresses link prediction and node classification simultaneously, the baseline models are customized for each task separately.} The VGAE+ is trained once for all subgraph queries, whereas the baseline models are trained for a specific task. Because the VGAE+ model is solving a harder problem, matching the performance of the baseline methods is a good result. Most baselines were developed for transductive graph prediction, some only transductively, some both transductively and inductively. We therefore expect them to perform relatively better in the semi-inductive than in the (fully) inductive setting. As we will see, our results confirm this expectation. 

\subsubsection{Subgraph Queries.} 
The reported AUC results for these baseline models are the average AUC scores for both link prediction and node classification tasks. We do not show standard deviations for simplicity, they range from 0.1 to 0.2. 
~\Cref{tab:joint} shows the results for subgraph queries, which predict {\em both} links and node labels. In the inductive setting, all target nodes, and potentially some of their neighbors, are unseen during training.

\begin{table*}[t]
\caption{AUC scores on Test Queries} \vskip 3mm
\begin{subtable}{1\textwidth}
\centering
\caption{AUC results for subgraph queries in semi-inductive and inductive settings.} 
\scalebox{0.9}{
\begin{tabular}{llrrrrrrrrrrrrr}\toprule
&&\multicolumn{6}{c}{\textbf{Single Neighbor Queries}} &&\multicolumn{6}{c}{\textbf{Neighborhood Queries}}\\
\cmidrule(lr){3-8} \cmidrule(lr){10-15}
           && Cora  & ACM & IMDb  & CiteSeer  & Photo & Computers && Cora  & ACM & IMDb  & CiteSeer  & Photo & Computers \\\midrule

\multicolumn{1}{c}{\multirow{4}{*}{\textbf{Semi-inductive}}}
& VGAE-Det& \underline{0.937} & 0.801 & 0.863 & \textbf{0.949} & \textbf{0.960} & 0.910 & & {0.868} & 0.700 & \textbf{0.764} & \underline{0.934} &0.854  & \underline{0.789} \\
& VGAE-MC &\textbf{0.949}  & \underline{0.828} & \underline{0.900} & \underline{0.941} & 0.951 & \textbf{0.960} & &\textbf{0.933} & \underline{0.716} & 0.734 & \textbf{0.951} & \textbf{0.949} & \textbf{0.918} \\
\cdashline{3-15}
& GAT  & 0.862& 0.909&0.843 &0.910 &0.935&0.922 &&0.829 &0.726 &0.733 &0.900 &0.881 & \underline{0.840} \\
& GraphSage  & 0.797 & 0.723 & 0.748 & 0.750 & 0.860 & 0.858 & &0.701 & 0.689 & 0.726 & 0.786 & 0.846 & 0.724 \\
 & GiGaMAE    & 0.913 & \textbf{0.883} & \textbf{0.920} & 0.896 & \underline{0.956} & \underline{0.948} & & \underline{0.917} & \textbf{0.782} & \underline{0.746} &  0.896 & \underline{0.900}& 0.703  \\
\midrule
\multicolumn{1}{c}{\multirow{5}{*}{\textbf{Inductive}}}
& VGAE-Det& \textbf{0.932} & 0.751 & \underline{0.880} & \textbf{0.963} & \underline{0.947} & \underline{0.945} & & 0.825& 0.697 & \underline{0.823} & \textbf{0.943} & \textbf{0.884}  & \textbf{0.885}    \\
& VGAE-MC & \underline{0.925} & \textbf{0.832} & \textbf{0.900} & \underline{0.912} & \textbf{0.953} & \textbf{0.954} & & \textbf{0.866} & \textbf{0.781} & \textbf{0.865} & \underline{0.889} & \underline{0.825} & \underline{0.819}\\
\cdashline{3-15}
& GAT  &0.728 &0.841 & 0.760& 0.798 & 0.849 & 0.866&& 0.681& 0.573& 0.650&0.648 & 0.630&0.582  \\
& GraphSage & 0.810 & \underline{0.783} & 0.644 & 0.702 & 0.761 & 0.712 &&0.710 & 0.475  & 0.463 & 0.709 & 0.684 & 0.566\\
 & GiGaMAE   & 0.896 & 0.668 & 0.715 & 0.896 & 0.788 & 0.805 & & \underline{0.828} & \underline{0.716} & 0.700 & 0.885 & 0.816 & 0.718
 \\\bottomrule
\end{tabular}
}
\label{tab:joint}
\end{subtable}

\hfill 
\vskip 2mm
\begin{subtable}{1 \textwidth}
    \centering
\caption{AUC results for link prediction in semi-inductive and inductive settings.} 

\scalebox{0.9}{
\begin{tabular}{llrrrrrrrrrrrrr}\toprule
&&\multicolumn{6}{c}{\textbf{Single Link Queries}} &&\multicolumn{6}{c}{\textbf{Joint Link Queries}}\\
\cmidrule(lr){3-8} \cmidrule(lr){10-15}
           && Cora  & ACM & IMDb  & CiteSeer  & Photo & Computers && Cora  & ACM & IMDb  & CiteSeer  & Photo & Computers \\\midrule

\multicolumn{1}{c}{\multirow{4}{*}{\textbf{Semi-inductive}}}
& VGAE-Det&  0.898& \underline{0.956}& 0.937 & 0.941 &0.947 & 0.871 & & {0.759} & \underline{0.962} & 0.894 & {0.910} & 0.798 & 0.753 \\
& VGAE-MC &\textbf{0.912}  & 0.954 & \underline{0.957} &\textbf{0.952}   & \textbf{0.956} & \textbf{0.935} && \underline{0.886} & 0.960 & 0.901 & \textbf{0.941}  & \textbf{0.940} &  \textbf{0.928} \\
\cdashline{3-15}
& SEAL    & \underline{0.910} & 0.904 & 0.906 & 0.948  & 0.935 & 0.908 & & {0.741} & 0.500 & 0.620 & 0.752   & 0.684 & 0.701   \\
& GAT   & 0.862& 0.943&0.874 &0.913 & 0.902 &0.892 & & 0.764 & 0.925 &0.845 & 0.904 & 0.872 & 0.814 \\
& GraphSage    & 0.854 & 0.969  & 0.917 & 0.589 & 0.858 &   0.913 & &   0.725  & \underline{0.962} &  0.893& 0.666 & 0.853 & 0.823 \\
 & GiGaMAE    & 0.907 & 0.943 & 0.950 &  \underline{0.950} & \underline{0.949} & \underline{0.917}  & & \textbf{0.902}& 0.959 & \textbf{0.963} & \underline{0.940} & \underline{0.930}  & \underline{0.906}  \\
 &DEAL & 0.800 & \textbf{0.986} & \textbf{0.981} & 0.914   & 0.832 & 0.823 & & 0.676 & \textbf{0.979} & \underline{0.962} & {0.910}  & {0.857} & {0.826}  \\
\midrule
\multicolumn{1}{c}{\multirow{5}{*}{\textbf{Inductive}}}
& VGAE-Det& \textbf{0.891} & 0.963 & 0.939 & \textbf{0.950} & \textbf{0.936} & \underline{0.925} && 0.690 & \underline{0.954} & \underline{0.867} & \textbf{0.925} & \textbf{0.857} & \textbf{0.821}   \\
& VGAE-MC & 0.865 & \textbf{0.970} & \underline{0.942} & 0.900 & \underline{0.931} & 0.922 & &\textbf{0.778} & \textbf{0.968} & \textbf{0.935} & 0.856 & 0.760 & 0.752 \\
\cdashline{3-15}
& SEAL   & 0.756 & 0.666 & 0.854 & 0.627  & 0.924   &  \textbf{0.972} && 0.693 & 0.681 & 0.830 & 0.679   & 0.506 & 0.450 \\
& GAT   & 0.667 & 0.870 & 0.796 & 0.731 &0.796 & 0.835 && 0.491 & 0.699 & 0.632 & 0.470 &0.453 &0.380  \\
& GraphSage  &   0.561     &      0.593  &   0.492    &     0.504   &    0.552    &  0.444 & &0.543    &     0.535   &  0.467    &    0.512     &  0.528  &     0.507 \\
 & GiGaMAE    & \underline{0.887} & 0.732 & 0.929 & \underline{0.943} & 0.798 & 0.764 & & \underline{0.725} & 0.920  & 0.918 & \underline{0.921} & \underline{0.854} & 0.706 \\
& DEAL   & 0.780 & \underline{0.902} & \textbf{0.957} & 0.861 & 0.852 & 0.844   & &  0.678 & 0.953 & \textbf{0.935} & 0.837 & 0.759 & \underline{0.757}
 \\\bottomrule
\end{tabular}}
\label{tab:lp}
\end{subtable}
\hfill
\vskip 2mm
\begin{subtable}{1 \textwidth}
\centering
\caption{AUC results for node classification in semi-inductive and inductive settings. 
} 
\scalebox{0.9}{
\begin{tabular}{llrrrrrrrrrrrrr}\toprule
&&\multicolumn{6}{c}{\textbf{Single Node Queries}} &&\multicolumn{6}{c}{\textbf{Joint Node Queries}}\\
\cmidrule(lr){3-8} \cmidrule(lr){10-15}
           && Cora  & ACM & IMDb  & CiteSeer  & Photo & Computers && Cora  & ACM & IMDb  & CiteSeer  & Photo & Computers \\\midrule

\multicolumn{1}{c}{\multirow{4}{*}{\textbf{Semi-inductive}}}
& VGAE-Det  & \underline{0.977} & 0.615 & 0.789  & \textbf{0.957} & \textbf{0.973} & 0.948 & & \underline{0.977} & 0.425  & \underline{0.634}  & \underline{0.959} & 0.910& 0.824    \\
& VGAE-MC & \textbf{0.986} & 0.701 & \underline{0.842} & \underline{0.930}  & 0.946& \textbf{0.986} && \textbf{0.980} & 0.471 & 0.568 & \textbf{0.960} & \textbf{0.959} &  \textbf{0.909} \\
\cdashline{3-15}
& GAT    &0.861 & 0.876 & 0.813&0.907 &0.967 &0.951 & & 0.893& 0.526& 0.621& 0.896&0.889 & 0.866\\
& GraphSage    & 0.740 & 0.477 & 0.580 & 0.911 & 0.863 & 0.804  & &0.677 &  0.416& 0.460 & 0.907 & 0.840 & 0.626 \\
 & GiGaMAE    & 0.920 & \textbf{0.823} & \textbf{0.890} & 0.842 & \underline{0.963} & 0.941 & & 0.932& \underline{0.604}  &0.529  &0.852  &0.871  &0.500  \\
 & MVGRL    & 0.888 & \underline{0.708} & 0.788 & 0.807 & \underline{0.963}  & \underline{0.980} & &   0.853 & \textbf{0.715} & \textbf{0.767} & 0.815 & \underline{0.953}  & \underline{0.892} \\
& Deep Walk     & 0.868& 0.643 &0.684&0.847&0.950&0.862&&0.726& 0.567&0.553&0.731&{0.937}&0.819   \\
\midrule
\multicolumn{1}{c}{\multirow{5}{*}{\textbf{Inductive}}}
& VGAE-Det  & \underline{0.974} & 0.540 & \underline{0.821} & \textbf{0.976} & 0.958 & 0.965 & & \textbf{0.961} & 0.441 & \underline{0.780} & \textbf{0.961} & \textbf{0.912} & \textbf{0.890}\\
& VGAE-MC  &\textbf{0.986}  &0.693  &  \textbf{0.856} & \underline{0.924} & \textbf{0.976} & \textbf{0.987} & & \underline{0.954} & \underline{0.595} & \textbf{0.796} & \underline{0.922} & 0.890 & \underline{0.886}  \\
\cdashline{3-15}
& GAT    &0.790 & 0.812 & 0.724 &0.865 &0.902 &0.896 & & 0.870 & 0.446& 0.668 & 0.826 &0.806 & 0.784\\
& GraphSage  & 0.970 & \textbf{0.973} & 0.797 & 0.900 & \underline{0.970} & \underline{0.980} & & 0.877& 0.416 & 0.460 & 0.907  & 0.840 & 0.626
\\
 & GiGaMAE     & 0.906 & 0.604 & 0.501 & 0.850 & 0.778 & 0.847 & & 0.932& 0.513 & 0.482 & 0.850 & 0.778 & 0.730 \\
 & MVGRL   & 0.650 & \underline{0.804} & 0.684 &  0.550&  0.835 & 0.801 & &   0.614 & \textbf{0.679} & 0.632 & 0.534 & \underline{0.892} & 0.827 \\\bottomrule
\end{tabular}}
\label{tab:nc}
\end{subtable}
\end{table*}

\paragraph{Neighborhood Queries} Considering the {\em inductive} setting, 
{\em both VGAE+ methods score more highly than both two-task baselines on almost all datasets}. The exceptions are the ACM and IMDb datasets, where the VGAE-Det score is similar to the GiGaMAE baseline, whereas VGAE-MC is clearly the best predictor. The biggest improvement for VGAE-MC is observed on the IMDb dataset (0.16 AUC points).  The strong performance on inductive queries illustrates how a generative model can generalize to data that may be incomplete or new, by inferring unseen information. 

In the {\em semi-inductive} setting, where some  target nodes are observed during training, VGAE-MC achieves the highest score on 4 out of 6 datasets, and a competitive score on IMDb. The VGAE-MC top scores are substantially higher than the baselines; for instance on Computers it outperforms by more than 0.2 AUC points. 

Our results show that for {\em some datasets, inference through sampling from an approximate posterior yields substantive accuracy improvements.} However, sampling takes longer to produce query results. To give a sense of the difference, for the biggest {\em Computers} dataset, sampling 30 node embeddings vs. using a single deterministic one takes about 4 times as long (39.02s/query for VGAE-MC vs. 11.951s/query for VGAE-Det), using a single NVIDIA A40 GPU.

\paragraph{Single Neighbor Queries} The task is to predict the link and node label for a single neighbor of a target node. VGAE-MC scores higher than both baselines {\em in the inductive setting}. The improvement over the next best baseline is substantive on 3 out of 6 datasets; for instance on IMDb, the improvement is almost 0.2 AUC points. 

GiGaMAE was designed for the transductive learning, and is generally competitive with VGAE-MC on single neighbor semi-inductive queries. VGAE-MC scores substantially higher than GraphSage, for instance on CiteSeer by almost 0.2 AUC points. 

In conclusion, {\em inference from a VGAE model offers the best performance for predicting jointly both links and node labels.}

We next break down the subgraph AUC scores into their link prediction and node label  components.
This allows us to understand the results in more detail and to compare with more single-task baselines.

\subsubsection{Link Prediction Queries.} ~\Cref{tab:lp} shows the AUC scores for link prediction queries. 
For {\em inductive} joint link prediction, VGAE-MC achieves top scores on 4 out 6 datasets compared to the baselines. GiGaMAE scores better on Photos and CiteSeer, although not as high as VGAE-Det inference. VGAE-MC outperforms the link prediction baselines other than DEAL and GiGaMAE. For {\em semi-inductive} joint link prediction, DEAL and GiGaMAE outperform VGAE-MC on 3 out of 6 datasets, but only by 0.02 AUC points at most (on Cora).

For {\em single-link inductive queries}, VGAE-MC outperforms the strongest baselines GiGaMAE and DEAL on 3 out of 6 datasets (ACM, Photo and Computers) by 0.08 AUC points or more. On the remaining 3 datasets, the VGAE-MC score is competitive. VGAE+ inference beats the other single-link prediction baselines, except for Photo and Computers, where SEAL performs exceptionally well. For {\em single-link semi-inductive queries} the VGAE-MC scores are the highest on 4 out of 6 datasets, but similar in magnitude to the GiGaMAE and DEAL baselines. 

In conclusion, {\em inference from a VGAE model offers excellent link prediction accuracy across a range of query tasks, including inductive and joint link prediction.}

\subsubsection{Node Classification Queries.} ~\Cref{tab:nc} shows AUC scores for node classification queries. For {\em inductive joint} node classification, VGAE-MC achieves a higher score than the baseline methods on 4 out of 6 datasets, especially on IMDb (at least 0.16 AUC points) and Cora (at least 0.07 AUC points). On ACM the strong node classification baseline MVGRL achieves a higher score. Similarly in the {\em semi-inductive joint} node classification setting, VGAE+ has a higher score than the baselines on 4 out of 6 datasets. The improvement is strongest on CiteSeer (0.06 over GraphSage). MVGRL is exceptionally strong in the semi-inductive setting on ACM and IMDb. 

For {\em inductive single} node classification, VGAE-MC again achieves the top score on 4 out of 6 datasets. GraphSage was designed for inductive node classification and is accordingly a strong baseline, especially on the ACM dataset. However, VGAE-MC outscores GraphSage by 0.06 AUC points on the IMDb dataset. 

For {\em semi-inductive single} node classification, VGAE-MC achieves the top score on 3 out of 6 datasets. The biggest improvement is observed on Cora, 0.06 AUC points over GiGaMAE. 

In conclusion, {\em inference from a VGAE model offers strong node classification accuracy across a range of settings, especially for joint node classification.}

\textbf{Subgraph Queries vs. Node Classification and Link Prediction.} Reviewing the results of inductive subgraph prediction in terms of link prediction and node classification, we observe that the high score of VGAE-MC on IMDb is due to its excellent score on  {\em joint node classification}. On the ACM dataset, VGAE-MC achieves strong  subgraph prediction through a high {\em joint  link prediction score}. The strong AUC score on Computers is due to both high joint node classification scores and high joint link prediction scores. 

Overall our experiments provide strong evidence that inference from an augmented VGAE model, {\em achieves an excellent balance between predicting links and node labels across different query types.} For single query types (e.g., link prediction), predictive performance is very competitive with custom baselines.

\textbf{Ablation Study.}~\Cref{tab:ablation} examines the importance of the components of our new training objective~\cref{eq:gen-loss} for the augmented VGAE. The second row ($\beta = 0$) shows that not reconstructing the node features leads to worse scores, especially on the IMDb and Computers datasets. This is remarkable since the neighborhood queries do not contain node features as a target. Turning off label reconstruction ($\gamma=0$) or link reconstruction ($\alpha =0$) leads to bad scores.

\begin{table}[ht]{
\caption{Ablation Study on the training objective~\cref{eq:gen-loss}.} \label{tab:ablation}

\scalebox{0.87}{
\begin{tabular}{llrrrrrrrrrrr}\toprule
&&\multicolumn{6}{c}{\textbf{Inductive Neighborhood Query}}
\\\cmidrule(lr){3-8}
           && Cora  & ACM & IMDb  & CiteSeer  & Photo & Computers\\\midrule
\multicolumn{1}{c}{\multirow{7}{*}}
& VGAE-Det &\textbf{0.825}& \textbf{0.697} & \textbf{0.823} & \textbf{0.943} & \textbf{0.884}  & \textbf{0.885}  \\
\multicolumn{1}{c}{} &  ($\beta=0$)& \underline{0.810} & 0.616 & 0.660& \underline{0.927}  & \underline{0.865} & \underline{0.748}\\
\multicolumn{1}{c}{} & ($\beta=0, \gamma=0$) & 0.611 & \underline{0.652} & \underline{0.710} & 0.688  & 0.653 &0.574\\
\multicolumn{1}{c}{} &  ($\beta=0, \alpha=0$)  & 0.656  & 0.417 &0.483 & 0.686  & 0.667 &0.580\\
\bottomrule
\end{tabular}}}
\end{table}

\section{Conclusion, Limitations, and Future Work} 

A subgraph prediction query asks for the probability of a target subgraph, given the information from an evidence subgraph. Supporting inference to answer subgraph queries (SQs) is a new use case for a deep Graph Generative Model (GGM). Such a query answering system facilitates applying graph prediction in a production environment where multiple users pose a range of queries to be answered. In this paper we showed how inference from a trained Variational Graph Auto-Encoder (VGAE) model, augmented with feature/label decoders, can be used to answer SQs, in zero-shot manner without retraining the model. Bayesian optimization was effective in balancing the relative importance of modeling links and node features/labels in a dataset-dependent manner. We carried out an empirical evaluation on six benchmark datasets and a range of test queries. The application of joint prediction from a single VGAE yielded higher accuracy than baseline methods that predict graph components independently. The strong performance of VGAE+ highlights the value of using both node features and graph structure as co-training objectives that improve both node classification and link prediction. 

{\em Limitations.} A limitation of our evaluation is that we considered only homogeneous graphs with a single link type. Deterministic and MC inference can be extended straightforwardly to knowledge graphs using a relational VGAE model. 

 {\em Future Work.} While the VGAE is a well-established GGM for a single training graph, other GGMs, especially auto-regressive and diffusion models, are known to have greater modeling power to capture complex correlation patterns in graphs~\cite{naddaf2023joint}. Leveraging the greater expressive power of these GGMs to improve subgraph predictions over our strong VGAE baseline is a fruitful direction for future research, especially if they can be trained on single graph inputs. Extending subgraph prediction to more complex graphs, such as weighted and/or dynamic graphs, is a fruitful topic for future work. 
Another valuable direction is to apply inference from a model to find the most likely subgraph given evidence. 

\clearpage

\bibliography{master}
\newpage
\section{Appendix Material.}
\subsection{Dataset Statistics}

To evaluate all the methods we utilize 6 datasets used in previous studies of generative models \cite{Kipf2016,yun2019GTN,hao2020inductive}.
Table \ref{tab:datasets} presents statistics for all 6 benchmark datasets. In this table \textit{edge density} refers to the number of links in a graph divided by the total number of possible links in that graph. This metric measures the connectedness or compactness of a graph. A lower number indicates a sparser graph, meaning there are fewer connections between the nodes. The \textit{average node degree} in a graph is the average of the degrees of all the nodes in the graph.

\begin{itemize}
    \item \textbf{Cora} is a citation dataset that consists of nodes that represent machine learning papers divided into seven classes. In this dataset nodes represent papers and links represent citations among them~\cite{sen2008collective}.
    \item \textbf{ACM} is a citation dataset. It has three types of nodes (author, paper, subject) and two types of edges (paper-author, paper-subject)~\cite{yun2019GTN}.
    \item \textbf{IMDb} is a movie dataset with three types of nodes (movie, actor, director) and two types of edges (movie-actor, movie-director). In this dataset labels are the genres of movies~\cite{yun2019GTN}.
    \item \textbf{CiteSeer} is also a citation dataset which consists of nodes that represent machine learning papers divided into six classes. In this dataset nodes represent papers and links represent citations among them~\cite{sen2008collective}.
    \item \textbf{Photo \& Computers} are datasets from the Amazon co-purchase graph~\cite{mcauley2015image}. In these datasets, nodes represent goods, links indicate that two goods are frequently bought together, node features are bag-of-words encoded product reviews, and class labels are given by the product category~\cite{hao2020inductive}.
    \end{itemize}

\begin{table}[ht]
\caption{Statistics of the datasets.}
\centering
\scalebox{0.9}{
  \begin{tabular}{lcccc}
    \toprule
     \textbf{Dataset} & \textbf{Nodes} & \textbf{Links} &  \textbf{Edge Density}  &  \textbf{Average Node Degree}
    \\
    \midrule
 Cora~\cite{Kipf2016} & 2,708 & 5,429  & 0.00074 & 3.8
   \\ 
ACM~\cite{yun2019GTN}& 8,993 & 18,929 & 0.00093 & 2.2
       \\
   IMDb~\cite{yun2019GTN} & 12,772 & 19,120  & 0.00046 & 2.9
   \\
      CiteSeer~\cite{Kipf2016} & 3,327 & 4,732 & 0.00171 & 2.7 
     \\
           Photo~\cite{hao2020inductive} & 7,650 & 238,162  & 0.01629  & 36.7
     \\
           Computers~\cite{hao2020inductive} & 13,752 & 491,722  &  0.01041  & 36.7
     \\
      
    \bottomrule
  \end{tabular}}
  \label{tab:datasets}
\end{table}

\subsection{Data Preprocessing}
Following previous link prediction studies~\citep{Kipf2016}, I add self-loops and make all links undirected (i.e., if the training data contains an adjacency, $\noden\rightarrow \nodem$, it also contains $\nodem\rightarrow \noden$). Cora, CiteSeer, Photo, and Computers are homogeneous datasets, whereas ACM and IMDb are heterogeneous datasets. Since our comparison methods use homogeneous GNNs, I homogenize different edge types so that every edge in the adjacency matrix is represented by 0 for no link and 1 for link existence.

\subsection{Evaluation}
\subsubsection{Metrics}
\paragraph{Hit-Rate@20\% (HR@20\%)}is computed as follows: (1) For each method, find the set $T$ the of top 20\% links as ranked by the method. (2) Find the number of ground-truth links in $T$. (3) Divide by the total number of links in $T$.
\paragraph{F1-score macro}is a metric used to evaluate the performance of a classification model, especially in situations where the dataset is imbalanced. It is the harmonic mean of precision and recall, calculated separately for each class, and then averaged. This ensures that each class is given equal weight, regardless of its prevalence in the dataset. F1-score macro is particularly useful because it highlights the performance of the model across all classes, preventing high performance on a dominant class from overshadowing poor performance on minority classes. This makes it a valuable metric for assessing models in multi-class and imbalanced data scenarios.
\subsubsection{Results}
\paragraph{Hit-Rate (HR)}
\cref{tab:hr} shows the performance of the VGAE+(MC) and VGAE+(Det) on HR@20\% metric. We Should consider that this metric is very sensitive to threshold that we set for the predicted probabilities by the model. For this metric, on \textbf{ single link inductive queries}, VGAE+(MC) outperforms or is equal to the strong baseline DEAL on 5 out of 6 datasets. On the Computers dataset, the VGAE+(MC) score is competitive. VGAE+ inference beats the other single link prediction baselines. For \textbf{ single link semi-inductive queries}, VGAE+(MC) outperforms or is equal to the strong baseline DEAL on 5 out of 6 datasets. On the IMDb dataset, the VGAE+(MC) score is competitive. On \textbf{ joint link inductive queries}, VGAE+(MC) outperforms or is equal to the strong baseline DEAL on 5 out of 6 datasets. On the CiteSeer dataset, the VGAE+(MC) score is competitive only 1\% less than DEAL. VGAE+ inference beats the other single link prediction baselines. For \textbf{ joint link semi-inductive queries}, VGAE+(MC) outperforms or is equal to the strong baseline DEAL on 4 out of 6 datasets. On the IMDb and CiteSeer datasets, the VGAE+(MC) score is only 2\% less than DEAL.
\paragraph{F1-score Macro}
\cref{tab:f1} demonstrate strong performance of VGAE+(MC) for f1-score macro metric. In both single node classification queries and joint node classification queries, we can see VGAE+(MC) is performing very well. For \textbf{inductive single} node classification, VGAE+(MC) achieves the top score on 5 out of 6 datasets, by almost 0.2 on IMDb. 
For \textbf{semi-inductive single} node classification, VGAE+ achieves the top score on 2 out of 6 datasets. VGAE+(Det) on Cora by 0.03 points and VGAE+(MC) on CiteSeer by 0.12 points. On the 4 remaining datasets, GAT outperforms on IMDb by 0.13 points, and MVGRL outperforms on Photo and Computers by at most 0.16 points.
in \textbf{inductive joint} node classification, VGAE+ achieves the top score on 4 out of 6 datasets. VGAE+(MC) outperforms on 2 datasets (Cora and Photo) by almost 0.08 points on Cora. VGAE+(Det) outperforms on 2 datasets (IMDb and CiteSeer) by almost 0.17 points on CiteSeer. MVGRL outperforms on ACM and Computers by 0.13 points.

For \textbf{semi-inductive joint} node classification, VGAE+ achieves the top score on 2 out of 6 datasets (VGAE+(MC) on Cora by 0.05 points and VGAE+(Det) on CiteSeer by 0.13 points), and is competitive on the remaining datasets. DeepWalk outperforms on IMDb by 0.33 points, and MVGRL outperforms on ACM, Photo, and Computers by at most 0.25 points on ACM.

\begin{table*}[t]
\caption{F1-score macro and Hit-Rate @20\% }
\vskip 3mm
\begin{subtable}{1\textwidth}
\centering
\caption{Hit-Rate @20\% results for both single link prediction and joint link prediction in  inductive settings}
\scalebox{0.9}{
\begin{tabular}{llrrrrrrrrrrrrr}\toprule
&&\multicolumn{6}{c}{\textbf{Single Link Queries}} &&\multicolumn{6}{c}{\textbf{Joint Link Queries}}\\
\cmidrule(lr){3-8} \cmidrule(lr){10-15}
           && Cora  & ACM & IMDb  & CiteSeer  & Photo & Computers && Cora  & ACM & IMDb  & CiteSeer  & Photo & Computers \\\midrule

\multicolumn{1}{c}{\multirow{4}{*}{\textbf{Semi-inductive}}}
& VGAE+(Det) & 87.5&\textbf{100} &\textbf{100} &95.9 &97.8 &\underline{92.5} & &83.9 &84.2 &84.7 & 91.6&\underline{84.6 } &85.3\\
& VGAE+(MC) & \textbf{100} & \textbf{100} & \underline{98.2}& \textbf{100} & \textbf{100}& \textbf{100} && \textbf{100}& \textbf{100}& \underline{98.5} & \underline{97.3} & \textbf{100} & \textbf{100} \\
\cdashline{3-15}
& GAT   &92.5 & \textbf{100}& \textbf{100} &95.0 & 94.2 & 91.5& &76.9 & 82.8& 78.9 & 92.0& 77.0& 72.8\\
& GraphSAGE    & 90.0& \textbf{100}&\textbf{100} & 92.5& \underline{97.5}& \underline{92.5}& &73.8 &\underline{98.2} &90.0 &94.7 & 83.1& \underline{85.5}\\
 & GiGaMAE    & \underline{96.5} & \underline{71.8} & 78.9 & \underline{97.5} & 85.0 &84.7 &&  \underline{94.6} & 69.9 & 72.5 & 95.8 & 76.7&84.1 \\
 &DEAL  &\textbf{100} & \textbf{100}& \textbf{100}& \textbf{100}& \textbf{100}& \textbf{100}  &&  \textbf{100} & \textbf{100}& \textbf{100}& \textbf{100}& \textbf{100}& \textbf{100}  \\
\midrule
\multicolumn{1}{c}{\multirow{5}{*}{\textbf{Inductive}}}
& VGAE+(Det) &87.5 &\textbf{100} & {95.0} &95.3 & 92.6 & 88.2  &&  63.9  &82.7 &80.4 & 90.5& \underline{86.7} &  85.3  \\
& VGAE+(MC) &\textbf{100} & \textbf{100}& \textbf{100} & \textbf{100} &\textbf{100} &{98.5}   &&  \textbf{100}  & \textbf{100}& \textbf{100}& \underline{98.6} & \textbf{100}  & \textbf{100} \\
\cdashline{3-15}
& SEAL   & \underline{97.9}& \underline{84.3}& 90.2& 97.4 & \underline{98.1} & \underline{99.5}  &&  52.0  & 11.5& 38.0 & 48.0 &4.00  &\textbf{100} \\
& GAT    &89.5 & \textbf{100}& \underline{98.8 }&92.1 & 96.4 & 90.0& &56.8 & 72.6& 64.5 & 81.0& 68.5& 65.8\\
& GraphSAGE  & 87.8& \textbf{100}& \textbf{100}&94.7 &93.7 & 96.1  && 68.8   & \underline{94.5}& \underline{87.8}& 90.5&  85.1&  78.5\\
 & GiGaMAE    & 95.0& 65.3& 54.2&\underline{97.5} &85.0 &85.0   && \underline{94.2}    &54.1 &30.6 &{96.8} & 70.6  & \underline{87.1}\\
& DEAL   &\textbf{100} & \textbf{100}& \textbf{100}& \textbf{100}& \textbf{100}& \textbf{100}  &&  \textbf{100} & \textbf{100}& \textbf{100}& \textbf{100}& \textbf{100}& \textbf{100}  \\\bottomrule
\end{tabular}}
\label{tab:hr}
\end{subtable}

\hfill

\begin{subtable}{1\textwidth}
\centering
\caption{F1-macro results for both single node classification and node classification in inductive settings}
\scalebox{0.9}{
\begin{tabular}{llrrrrrrrrrrrrr}\toprule
&&\multicolumn{6}{c}{\textbf{Single Node Queries}} &&\multicolumn{6}{c}{\textbf{Joint Node Queries}}\\
\cmidrule(lr){3-8} \cmidrule(lr){10-15}
           && Cora  & ACM & IMDb  & CiteSeer  & Photo & Computers && Cora  & ACM & IMDb  & CiteSeer  & Photo & Computers \\\midrule

\multicolumn{1}{c}{\multirow{4}{*}{\textbf{Semi-inductive}}}
& VGAE+(Det)   & 0.792 & 0.498 & 0.487 &\textbf{0.831}& 0.742 & 0.687 && \underline{0.872}   &{0.411} & 0.451 &\textbf{0.914} &0.698  & 0.638 \\
& VGAE+(MC)  & \textbf{0.894} &  0.606 & 0.548 & 0.791 &0.801 & \underline{0.758} & &\textbf{0.903} & 0.450 & 0.448  &\underline{ 0.816}& 0.758& \underline{0.715}\\
\cdashline{3-15}
& GAT     & 0.529 &0.570 &\textbf{0.671} &0.709 &0.814 &0.471 & &0.644 &0.468 &0.426 &0.658 & 0.606& 0.488\\
& GraphSAGE   &0.551 & 0.416& 0.439&0.779 &0.598 &0.399 &&0.486 &0.435 &0.475 &0.777 &0.476 &0.362 \\
& GiGaMAE    &0.856 & 0.440 &0.457 & \underline{0.798}& 0.770& 0.569&& 0.784 &0.430 &0.451 &0.748 &0.797 &0.490 \\
& MVGRL  & \underline{0.886} & \textbf{0.803} &\underline{ 0.653} & 0.710 & \textbf{0.960} & \textbf{0.816} && 0.789 & \textbf{0.708} & \underline{0.604} & 0.717 &\textbf{ 0.950} & \textbf{0.739}\\
& DeepWalk  &0.700 &\underline{0.632} &0.650 & 0.700&\underline{0.850} &0.750 && 0.856&\underline{0.605 }&\textbf{0.787} &0.632 &\underline{0.851} &0.663 \\
\midrule
\multicolumn{1}{c}{\multirow{5}{*}{\textbf{Inductive}}}
& VGAE+(Det)  & 0.760&0.411 & 0.500 &0.950 & 0.785& 0.643  && \underline{0.863}   &\underline{0.594} & \textbf{0.542}&\textbf{0.905} &0.628  &0.534 \\
& VGAE+(MC)  & \textbf{0.855} &0.372 &\textbf{0.741} & \textbf{0.794} & \textbf{0.863}& \textbf{0.758}  &&  \textbf{0.935}  &0.442 & \underline{0.481} & \underline{0.791}& \textbf{0.845} & \underline{0.674}\\
\cdashline{3-15}
& GAT    & 0.628 &0.543 &0.470 &0.665 &0.724 &0.498 & &0.540 &0.448 &0.461 &0.435 & 0.586& 0.434\\
& GraphSAGE &0.486 &\underline{0.416} &0.439 &0.777 &\underline{0.850} &  0.362 && 0.351   & 0.335&0.437 &0.771 & 0.789 & 0.393
\\
 & GiGaMAE    & \underline{0.847}&0.406&0.441 &\underline{0.786} &0.779 &0.340    &&  0.852  & 0.189&0.421 & 0.732& 0.793 & 0.320 \\
 & MVGRL  & 0.430&\textbf{0.786} & \underline{0.460} & 0.416& 0.743&\underline{0.720}   && 0.501   & \textbf{0.735} & 0.478&0.365 & \underline{0.801} &\textbf{0.706}  \\\bottomrule
\end{tabular}}
\label{tab:f1}
\end{subtable}
\end{table*}
\begin{figure}[ht]
\centering
    \includegraphics[width=\columnwidth]
    {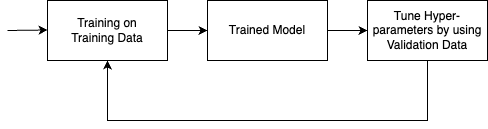}
    \caption{Hyper-parameters tuning process\\}
    \label{fig:tuning}
\end{figure} 
\subsection{Bayesian optimization}
 Bayesian optimization is a sequential model-based optimization technique used for optimization. It is an algorithm for finding the best set of parameters for a given objective function. Bayesian optimization efficiently searches the parameter space and often finds optimal or near-optimal solutions with fewer function evaluations compared to grid search or random search methods by intelligently selecting points to evaluate.

\subsubsection{Tuning Hyper-parameters}
The weight hyperparameters $\alpha$, $\beta$, $\gamma$ are found by Bayesian optimization~\cite{bib:bayesopt}. The higher value for each of them shows the greater effect of the correspondence reconstruction term. By tuning these hyperparameters VGAE+ is trained in a dataset-dependent manner.
The optimizer optimizes these parameters over the interval [0,1] so that they minimize the value of the reconstruction loss:

\begin{equation}
\begin{split}
min_{\alpha,\beta,\gamma}E_{\Z \sim q_{\efpars}(\Z|\Xvariable,\A)}[
  \ln{ p_{\aparameters}(\A|\Z})
+ \ln{ p_{\pparameters}(\X|\Z})
+ \ln{ p_{\efpars}(L|\Z})\big] 
\\
\end{split}
\label{eq:bayesian_Opt}
\end{equation}
\Cref{fig:tuning} shows the process of Hyper-parameters tuning. 

\subsection{Posterior Distribution $p(\Z|\ev = \evinstance)$}

\begin{figure}
  \centering
      \includegraphics[width=0.4\textwidth]{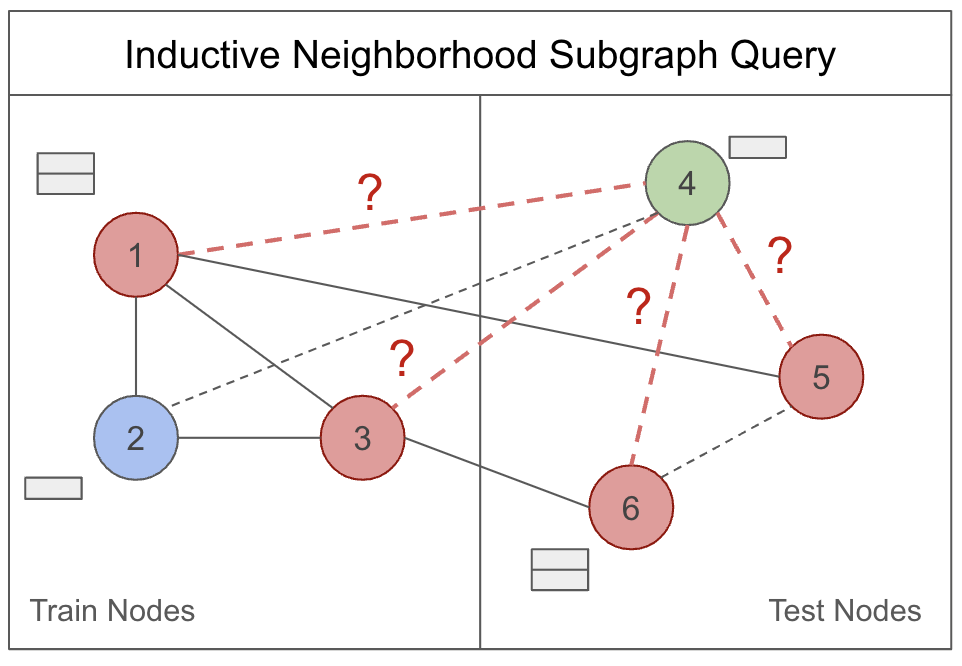}
  \caption{An example of inductive subgraph query: Red dashed links are target links, and black dashed links are unspecified links. Red colored nodes are target nodes for node classification. The squares beside each node represent features for that node.}
  \label{fig:sq}
\end{figure}

To illustrate how we handle unspecified links in the evidence, consider Figure \ref{fig:sq}. Since the link between nodes 1 and 2 is specified to exist, the evidence adjacency matrix assigns $\A^{\ev,\set{0}}[1,2] := 1$. The link between nodes 3 and 5 is specified not to exist, so the evidence matrix assigns $\A^{\ev,\set{0}}[3,5] := 0$. 
Since the link between nodes 5 and 6 is unspecified, the evidence matrix assigns $\A^{\ev,\set{0}}[5,6] := 0$. 
The feature vector for node 3 is unspecified, the evidence matrix assigns the zero feature vector $\Xvariable^{\ev,\set{0}}[2] := \set{0}$. 
The following matrices demonstrate $\evidence{\Avars}$, $\A^{\ev,\set{0}}$, $\evidence{\Xvars}$, and $\Xvariable^{\ev,\set{0}}$ for the example query in Figure \ref{fig:sq}, where links (2, 4) and (5, 6) and features for nodes 3, 5 are unspecified. Unspecified values are shown with ``?'':

\begin{itemize}
    \item $\evidence{\Avars} = \begin{bmatrix}
1 & 1 & 1 & 0 & 1 & 0\\
1 & 1 & 1 & ? & 0 & 0\\
1 & 1 & 1 & 0 & 0 & 1\\
1 & ? & 0 & 1 & 0 & 0\\
1 & 0 & 0 & 0 & 1 & ?\\
0 & 0 & 1 & 0 & ? & 1
\end{bmatrix}$

    \item $\A^{\ev,\set{0}} = \begin{bmatrix}
1 & 1 & 1 & 0 & 1 & 0\\
1 & 1 & 1 & 0 & 0 & 0\\
1 & 1 & 1 & 0 & 0 & 1\\
1 & 0 & 0 & 1 & 0 & 0\\
1 & 0 & 0 & 0 & 1 & 0\\
0 & 0 & 1 & 0 & 0 & 1
\end{bmatrix}$

    \item $\evidence{\Xvars} = \begin{bmatrix}
1 & 1 \\
0 & 1 \\
? & ? \\
1 & 0 \\
? & ? \\
1 & 1 
\end{bmatrix}$

    \item $\Xvariable^{\ev,\set{0}} = \begin{bmatrix}
1 & 1 \\
0 & 1 \\
0 & 0 \\
1 & 0 \\
0 & 0 \\
1 & 1 
\end{bmatrix}$

\end{itemize}

\subsection{Subgraph Query Examples}
\begin{figure}[ht]
\centering
    \includegraphics[width=\columnwidth]
    {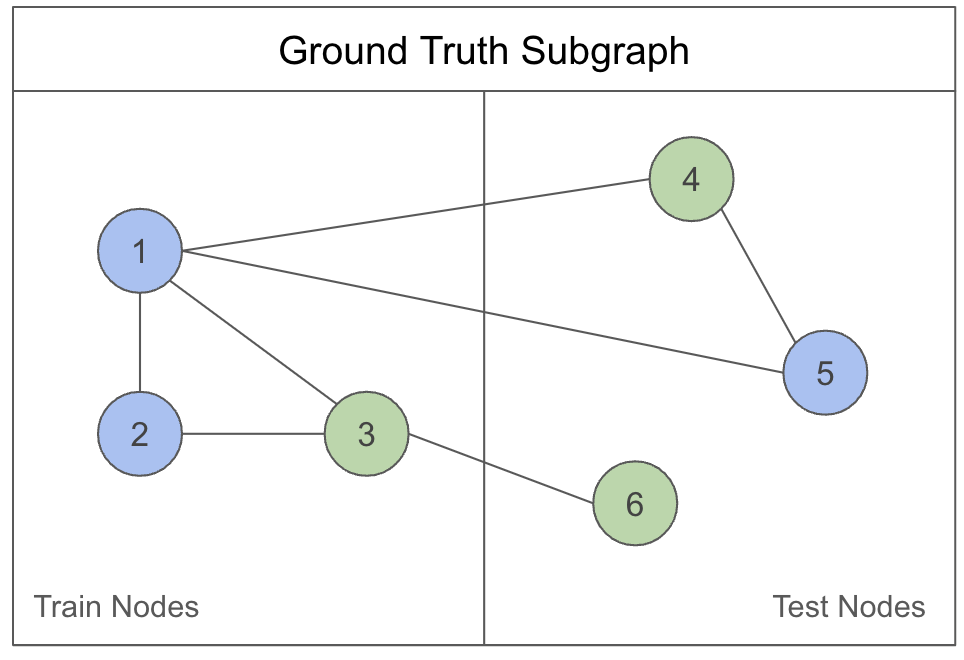}
    \caption{Input graph with partition of nodes. Node labels are green and blue.}
    \label{fig:input_graph_appendix}
\end{figure}

\begin{figure}[ht]
\centering
    \begin{subfigure}[b]{0.23\textwidth}
        \includegraphics[width=\textwidth]{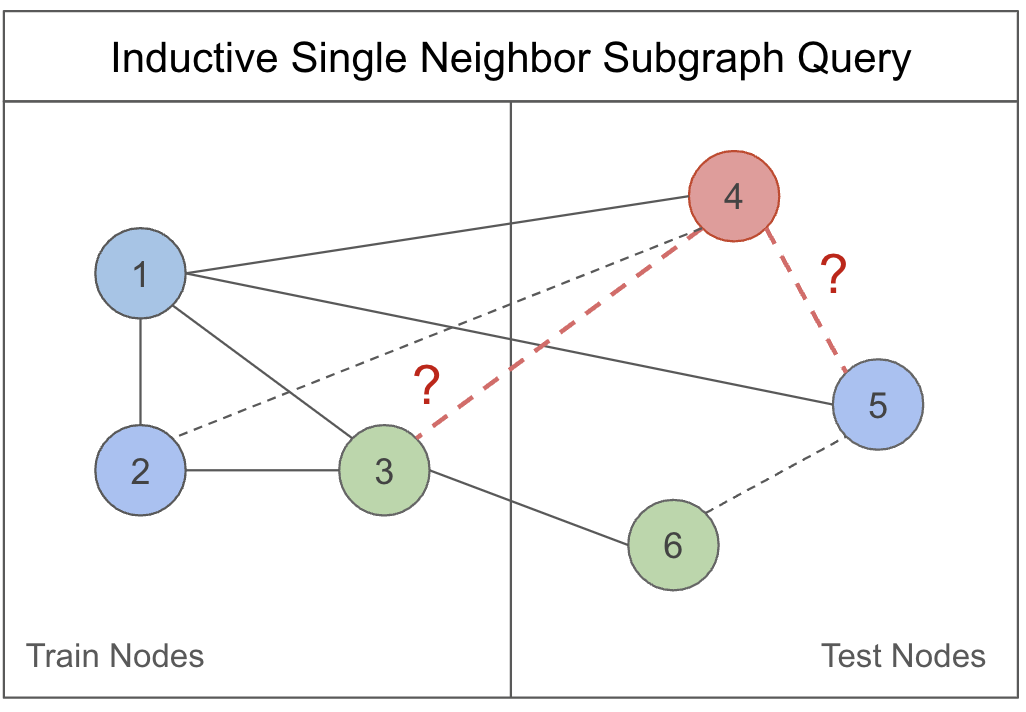}
    \caption{Single Neighbor query}
    \label{fig:neighbor}
    \end{subfigure}
    \begin{subfigure}[b]{0.23\textwidth}
        \includegraphics[width=\textwidth]{figures/Neighborhood_query.png}
    \caption{Neighborhood query}
    \label{fig:neighborhood}
    \end{subfigure}
    \caption{Examples of inductive subgraph queries. Target links are colored red. Black dashed links are unspecified links in evidence.}
    \label{fig:subgraph}
\end{figure}

\cref{fig:input_graph_appendix} is the ground truth that we expect from our model to predict. 
\paragraph{Single Neighbour Query}
In \cref{fig:neighbor} you can see the Single neighbour setting where:
\begin{itemize}
    \item The query nodes are 1--6.
    \item The {\em query target subgraph} comprises:
    \begin{itemize}
    \item one positive target link defined by the pairs $(5,4)$.
    \item one negative target link defined by the pairs $(3,4)$. 
        \item the labels of nodes $4$, 
    \end{itemize}
    \item The {\em query evidence subgraph} comprises:
    \begin{itemize}
       \item six positive evidence links defined by the pairs $(1,2),(1,3),(1,4),(2,3),(3,6),(1,5)$. 
    \item five negative evidence links defined by the pairs $(3,5),(2,6),(4,6),(2,5),(1,6)$. 
    \item The features of all query nodes (not shown). 
    \end{itemize}
    \item The presence/absence of links $(2,4)$ and $(6,4)$ is unspecified. 
\end{itemize}


\paragraph{Neighborhood Query} \cref{fig:neighborhood} shows a neighborhood query where:
\begin{itemize}
    \item The query nodes are 1--6.
    \item The {\em query target subgraph} comprises:
    \begin{itemize}
    \item two positive target links defined by the pairs $(1,4),(5,4)$.
    \item two negative target links defined by the pairs $(3,4),(6,4)$. 
        \item the labels of nodes $1,3,5,6$, 
    \end{itemize}
    \item The {\em query evidence subgraph}comprises:
    \begin{itemize}
       \item five positive evidence links defined by the pairs $(1,2),(1,3),(2,3),(3,6),(1,5)$. 
    \item four negative evidence links defined by the pairs $(3,5),(2,6),(2,5),(1,6)$. 
    \item The features of all query nodes (not shown). 
    \end{itemize}
    \item The presence/absence of links $(2,4)$ and $(6,4)$ is unspecified. 
\end{itemize}

\paragraph{Single Link Query} \cref{fig:neighborhood} shows a single link query where:
\begin{itemize}
    \item The query nodes are 1--6.
    \item The {\em query target set} comprises:
    \begin{itemize}
    \item one positive target link defined by the pairs $(5,4)$.
    \item one negative target link defined by the pairs $(3,4)$.  
    \end{itemize}
    \item The {\em query evidence set} comprises:
    \begin{itemize}
       \item six positive evidence links defined by the pairs $(1,2),(1,3),(1,4),(2,3),(3,6),(1,5)$. 
    \item five negative evidence links defined by the pairs $(3,5),(2,6),(4,6),(2,5),(1,6)$. 
    \item The features of all query nodes (not shown). 
    \end{itemize}
    \item The presence/absence of links $(2,4)$ and $(6,4)$ is unspecified. 
\end{itemize}

\paragraph{Joint Link Query} \cref{fig:neighborhood} shows a joint link query where:
\begin{itemize}
    \item The query nodes are 1--6.
    \item The {\em query target set} comprises:
    \begin{itemize}
    \item two positive target links defined by the pairs $(1,4),(5,4)$.
    \item two negative target links defined by the pairs $(3,4),(6,4)$.  
    \end{itemize}
    \item The {\em query evidence set} comprises:
    \begin{itemize}
       \item five positive evidence links defined by the pairs $(1,2),(1,3),(2,3),(3,6),(1,5)$. 
    \item four negative evidence links defined by the pairs $(3,5),(2,6),(2,5),(1,6)$. 
    \item The features of all query nodes (not shown). 
    \end{itemize}
    \item The presence/absence of links $(2,4)$ and $(6,4)$ is unspecified. 
\end{itemize}

\paragraph{Single Node Query}
In \cref{fig:neighbor} you can see the Single node query setting where:
\begin{itemize}
    \item The query nodes are 1--6.
    \item The {\em query target set} comprises:
    \begin{itemize}

    \item the labels of nodes $4$, 
    \end{itemize}
    \item The {\em query evidence set} comprises:
    \begin{itemize}
       \item seven positive evidence links defined by the pairs $(1,2),(1,3),(1,4),(2,3),(3,6),(1,5),(4,5)$. 
    \item six negative evidence links defined by the pairs $(3,5),(2,6),(3,4),(4,6),(2,5),(1,6)$. 
    \item The features of all query nodes (not shown). 
    \end{itemize}
    \item The presence/absence of links $(2,4)$ and $(6,4)$ is unspecified. 
\end{itemize}

\paragraph{Joint Node Query} \cref{fig:neighborhood} shows a Joint node query where:
\begin{itemize}
    \item The query nodes are 1--6.
    \item The {\em query target set} comprises:
    \begin{itemize}
        \item the labels of nodes $1,3,5,6$, 
    \end{itemize}
    \item The {\em query evidence subgraph}comprises:
    \begin{itemize}
       \item seven positive evidence links defined by the pairs $(1,2),(1,3),(1,4),(2,3),(3,6),(1,5),(4,5)$. 
    \item six negative evidence links defined by the pairs $(3,5),(2,6),(3,4),(4,6),(2,5),(1,6)$.  
    \item The features of all query nodes (not shown). 
    \end{itemize}
    \item The presence/absence of links $(2,4)$ and $(6,4)$ is unspecified. 
\end{itemize}

\begin{figure}[ht]
\centering
    \begin{subfigure}[b]{0.23\textwidth}
        \includegraphics[width=\textwidth]{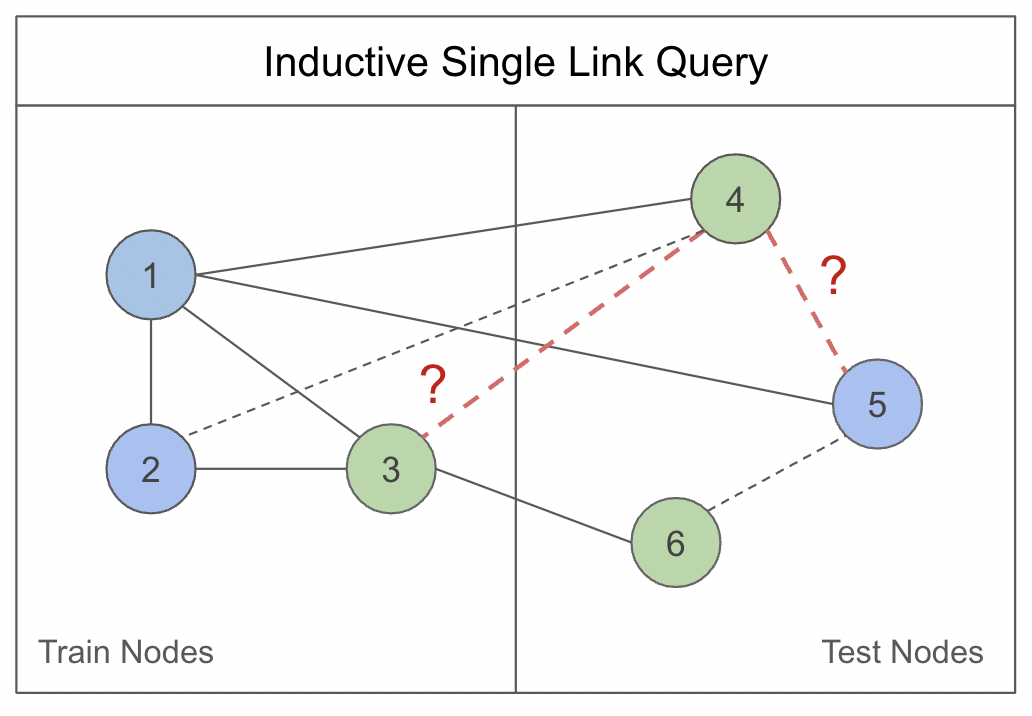}
    \caption{Single link query}
    \label{fig:slq}
    \end{subfigure}
    \begin{subfigure}[b]{0.23\textwidth}
        \includegraphics[width=\textwidth]{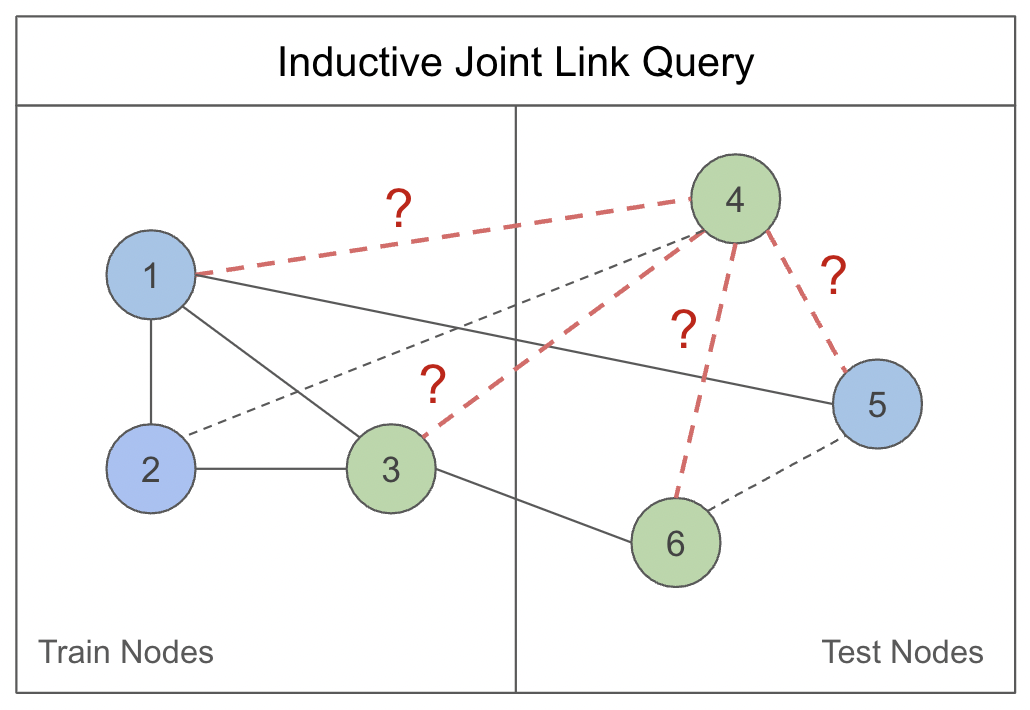}
    \caption{Joint link query}
    \label{fig:jlp}
    \end{subfigure}
    \caption{Examples of inductive link queries. Target links are colored red. Black dashed links are unspecified links in evidence.}
    \label{fig:lq}
\end{figure}

\begin{figure}[ht]
\centering
    \begin{subfigure}[b]{0.23\textwidth}
        \includegraphics[width=\textwidth]{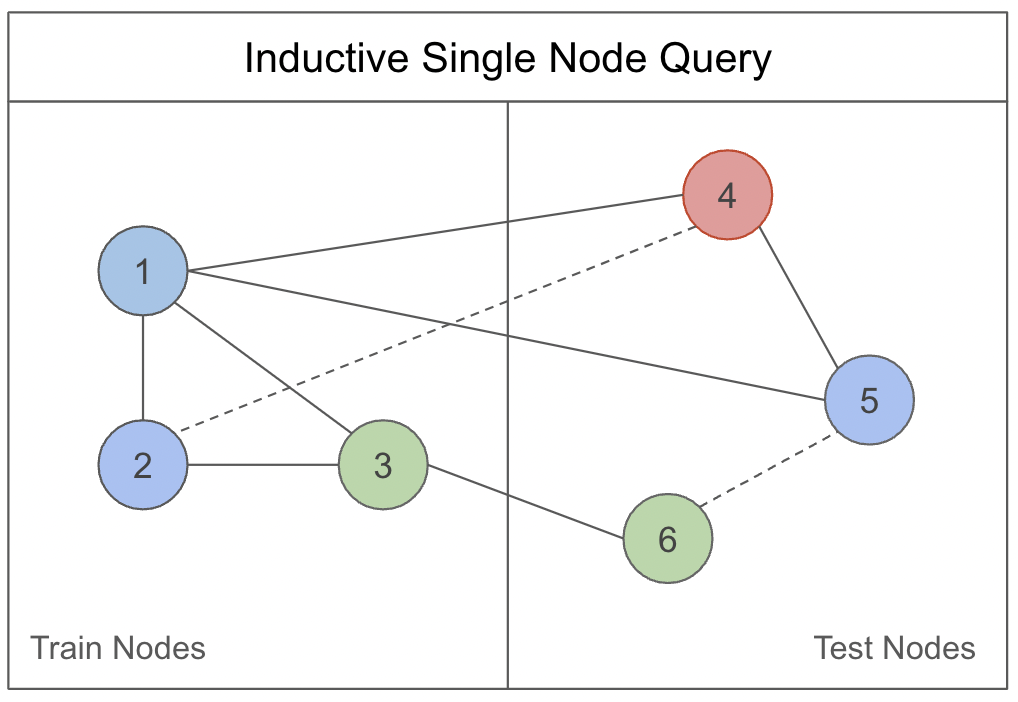}
    \caption{Single node query}
    \label{fig:snc}
    \end{subfigure}
    \begin{subfigure}[b]{0.23\textwidth}
        \includegraphics[width=\textwidth]{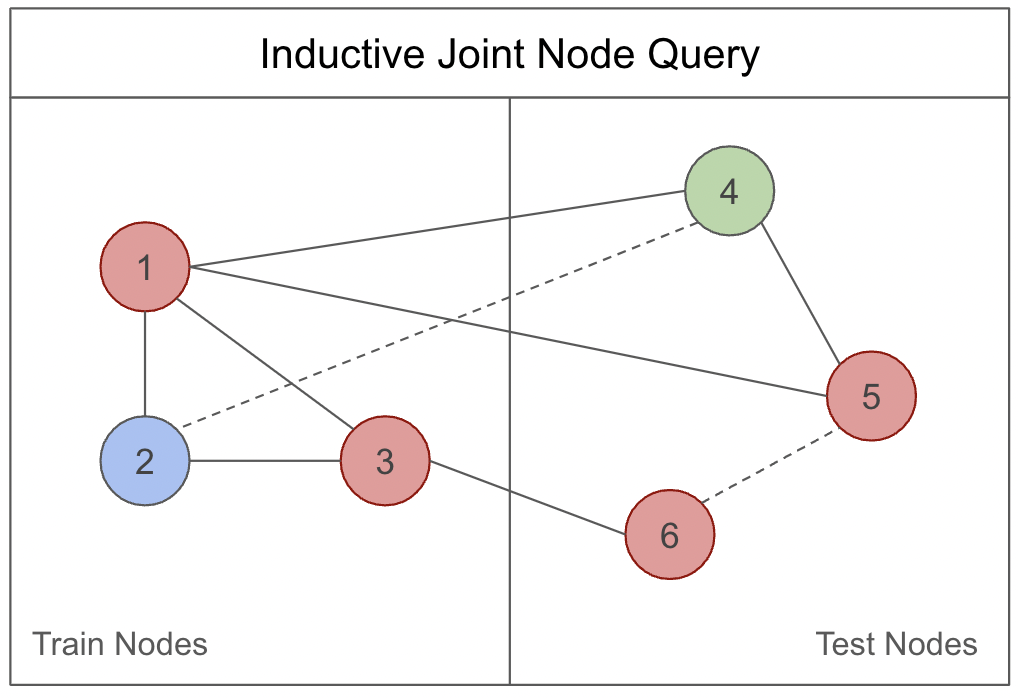}
    \caption{Joint node query}
    \label{fig:jnc}
    \end{subfigure}
    \caption{Examples of inductive node queries. Target links are colored red. Black dashed links are unspecified links in evidence.}
    \label{fig:nc}
\end{figure}



\end{document}